\documentclass[a4paper,fleqn]{cas-dc}

\usepackage[square,numbers,sort&compress]{natbib}
\usepackage{algorithm}
\usepackage{algorithmic}
\usepackage{graphicx}
\usepackage{textcomp}
\usepackage{multirow}
\usepackage{xcolor}
\usepackage{stfloats}
\usepackage{booktabs} 
\usepackage[T1]{fontenc} 
\usepackage[utf8]{inputenc}
\usepackage{threeparttable}
\usepackage{subfigure}
\usepackage{url}
\usepackage{xcolor}
\usepackage{verbatim}
\usepackage{color}
\usepackage{diagbox}
\usepackage{threeparttable}
\usepackage{caption}

\newcommand{\algrule}[1][.2pt]{\par\vskip.5\baselineskip\hrule height #1\par\vskip.5\baselineskip}
\newcommand{\SWITCH}[1]{\STATE \textbf{Switch} (#1)}
\newcommand{\ENDSWITCH}{\STATE \textbf{end switch}}
\newcommand{\CASE}[1]{\STATE \quad \textbf{case} #1\textbf{:} \begin{ALC@g}}
\newcommand{\ENDCASE}{\end{ALC@g}}
\newcommand{\CASELINE}[1]{\STATE \quad \textbf{case} #1\textbf{:} }

\def\tsc#1{\csdef{#1}{\textsc{\lowercase{#1}}\xspace}}
\tsc{WGM}
\tsc{QE}
\tsc{EP}
\tsc{PMS}
\tsc{BEC}
\tsc{DE}

\begin{document}
\let\WriteBookmarks\relax
\def\floatpagepagefraction{1}
\def\textpagefraction{.001}

\shorttitle{Non-negative Subspace Feature Representation for Few-shot Learning in Medical Imaging}

\shortauthors{Keqiang Fan et~al.}

\title [mode = title]{Non-negative Subspace Feature Representation for Few-shot Learning in Medical Imaging}                      



%
\author[1]{Keqiang Fan}[
                        style=chinese,
                        orcid=0000-0002-9411-2892
                        ]

\cormark[1]


\ead{kf1d20@soton.ac.uk}



\affiliation[1]{organization={Electronics and Computer Science, University of Southampton},
    city={ Southampton},
    postcode={SO17 1BJ}, 
    country={UK}}

\author[1]{Xiaohao Cai}[style=chinese,
                        orcid=0000-0003-0924-2834]
\ead{x.cai@soton.ac.uk}
\author[1]{Mahesan Niranjan}[%
    orcid=0000-0001-7021-140X]
\ead{mn@ecs.soton.ac.uk}

\cortext[cor1]{Corresponding author}

\begin{abstract}
Unlike typical visual scene recognition domains, in which massive datasets are accessible to deep neural networks, medical image interpretations are often obstructed by the paucity of data.
In this paper, we investigate the effectiveness of data-based few-shot learning in medical imaging by exploring different data attribute representations in a low-dimensional space. We introduce different types of non-negative matrix factorization (NMF) in few-shot learning, addressing the data scarcity issue in medical image classification. Extensive empirical studies are conducted in terms of validating the effectiveness of NMF, especially its supervised variants (e.g., discriminative NMF, and supervised and constrained NMF with sparseness), and the comparison with principal component analysis (PCA), i.e., the collaborative representation-based dimensionality reduction technique derived from eigenvectors.
With $14$ different datasets covering $11$ distinct illness categories, thorough experimental results and comparison with related techniques demonstrate  that NMF is a competitive alternative to PCA  for few-shot learning in medical imaging, and the supervised NMF algorithms are more discriminative in the subspace with greater effectiveness. 
Furthermore, we show that the part-based representation of NMF, especially its supervised variants, is dramatically impactful in detecting lesion areas in medical imaging with limited samples.
\end{abstract}












\maketitle
\section{Introduction}\label{Introduction}
%
 Recent remarkable advancements in computer vision have prompted an interest in expanding these technologies for diagnostic and prognostic inference based on medical images. 
 In medical imaging, the range of manually extracted image features is often limited. For different diseases, extracting relevant features from complex objects such as lesions and organs can be a challenging task due to their intricate nature, or may be relatively straightforward to achieve. It is advantageous to utilize neural networks, as they provide a consensus way to resolve feature extraction capabilities in medical imaging.
 According to neural scaling laws \cite{Kaplan2020ScalingLF}, the performance of a neural network can be improved consistently given the increase of three factors: model size, dataset size, and the amount of computation engaged in training, illustrating the importance of complex parametric models and large-scale datasets in benefiting the performance of neural networks in computer vision applications. Therefore, research for imaging inference with networks is frequently motivated by very large neural network architectures (especially depth) and equally enormous datasets to address the inability of simple models in parameter estimation of overly complex objects in images.

In contrast to natural images, deep learning techniques in medical imaging encounter multiple challenges \cite{razzak2018deep}. One of the biggest challenges is data scarcity, i.e., the number of images available in the medical field is generally several orders of magnitude lower than that in many other fields; moreover, data disparities are considerable between different tasks in medical imaging. 
For instance, participants in the dataset used in the "Leipzig Study for Mind-Body-Emotion Interactions" (LEMON) \cite{babayan2019mind} represent only 10\% of those included in DeepLesion \cite{yan2017deeplesion}, which stands as the largest open dataset of clinical CT scans sourced from the NIH Clinical Center.
Due to constraints such as expert annotation costs and privacy concerns, constructing large and diverse enough medical datasets for deep learning models, such as disease detection, is extremely difficult. 
Although the combination of clinical information and medical images prompts several researchers to achieve improvements e.g. by exploring causality \cite{castro2020causality} and uncertainty \cite{laves2020well}, the fundamental problem of data scarcity in the medical field is again critical but has not been addressed well.

This paper focuses on medical image classification, particularly in the low data regime.
Our interest lies in exploring datasets even smaller than those typically required for the general supervised training paradigm, ranging from a few hundred to even just a few tens of images for each disease.
In this case, training a deep learning model from scratch is infeasible due to problems such as over-fitting and poor generalisation.
Transfer learning, while fine-tuning the networks' parameters pre-trained from e.g. natural images could be useful to some extent to alleviate the data scarcity issue in the target domain \cite{weiss2016survey}. However, for medical images, it does not considerably enhance classification performance \cite{raghu2019transfusion}.
Though more recent work using cascade learning (i.e., layer-wise pre-training of source models) \cite{wang2022deep} attempts to overcome this limitation, the work in \cite{raghu2019transfusion} found that the effects of transfer learning on two large-scale medical imaging datasets (retinal fundus and chest X-ray data) are mainly because of model over-parameterization rather than complicated feature reuse, as widely believed. 
While alternative methods such as data augmentation and data synthesis expand data  and add features, they may struggle to address the inherent bias of datasets lacking diversity compared to test data, potentially introducing noise, as it is  assumed that both training and test data are drawn from the same distribution \cite{shorten2019survey}.

``Few-shot learning" techniques have been proven to be quite effective to address the data scarcity challenge \cite{wang2020generalizing}. In contrast to current few-shot learning methods based on models and prior information for parameter adjustment {(e.g., Reptile \cite{nichol2018reptile}, MAML \cite{finn2017model}, prototypical network \cite{snell2017prototypical}, and matching network \cite{vinyals2016matching})}, data-based few-shot learning methods can leverage traditional machine learning techniques. Traditional machine learning models focus more on manual prior information, including but not limited to data cleaning, data preprocessing, feature extraction, feature intersection, etc. 
Influenced by data attributes on pattern classification \cite{xu2019sparse}, we intend to explore problems in medical scenarios from the characteristics of the data, e.g., the output from the penultimate layer of a pre-trained network. This will allow us to explore data representation in different subspaces that are helpful for pattern classification.

When analysing data in different spaces, the data quantity  will be different from the dimension of the one extracted by pre-trained deep neural networks. Since the limited parameters in traditional machine learning models could pose challenges in representing data with high dimensions, it is crucial to improve the robustness in subspaces and avoid the ``curse of dimensionality” with appropriate data representation techniques (including e.g. sparse, collaborative, and non-negative representation), especially in the few-shot learning scenario. For example, effective dimensionality reduction methods can reduce model complexity and enhance generalisation performance by preserving specific characteristics and eliminating collinearity between data attributes.

One dimension reduction method for maintaining data collaboration is principal component analysis (PCA) \cite{papailiopoulos2013sparse}, implemented by singular value decomposition (SVD) \cite{raghu2017svcca}. It is one of the most popular dimensionality reduction techniques and has been widely used to explore subspace representations \cite{simon2020adaptive,chan2015pcanet}.
However, a fundamental weakness of PCA/SVD is that its variance-preserving low-rank approximation properties are mainly suitable for unimodal and Gaussian-distributed data.
In the case of classification problems, the feature space is generally multimodal. Therefore, the representation of the principal components can have a certain ambiguity among the classes since the principal components with small eigenvalues may also contain critical divergence information. This implies, in some cases, the results using PCA for dimensionality reduction may not be as interpretable as just using the original sample features.

Pattern classification based on sparse representation and non-negative representation has been widely studied in tasks such as face recognition and object classification \cite{xu2019sparse}. The work in \cite{belhumeur1997eigenfaces} discovered that the effective representation of homogeneous data samples should be dense and non-negative, which is linked to non-negative matrix factorization (NMF) \cite{lee1999learning}. In contrast to PCA, there is no subtraction during NMF-based reconstruction, and its non-negative constraint promotes the intuitive notion of combining parts into a whole -- the part-based representation. 
The representation power of homogeneous samples can be boosted with the non-negative property while constraining heterogeneous samples, making the representation sparse and discriminative simultaneously and providing an efficient direction in the subspace for classification.

In this paper, different from the traditional paradigms of few-shot learning, we focus on the data-based few-shot learning paradigm, specifically targeting the utilization of subspaces in medical imaging. 
We investigate the intrinsic data representation within the features extracted by a pre-trained model, leveraging the benefits of information preservation in the subspace to address challenges in scenarios where the feature dimension exceeds the magnitude of available data.
Our main objective is for the first time to explore the innovative application of NMF, particularly its discriminant variants, as alternatives to SVD for dimensionality reduction in low data regimes for multiclass medical inference problems.
As a matrix factorization methodology, NMF generates sparse subspaces with part-based representations. Under the premise of sparsity and non-negativity, supervised NMF approaches such as discriminative NMF (DNMF) \cite{babaee2016discriminative} and supervised and constrained NMF with sparseness (SCNMFS) \cite{cai2018supervised} are also investigated for enhancing the subspace discriminative property by combining labelled samples. Varied combinations of the labelled samples will have different impacts on the subspace throughout the decomposition process. 
We demonstrate the robustness and generalizability of our method with these subspace representations across 14 datasets, covering 11 distinct medical classification tasks spanning four different imaging modalities. 
Thorough experimental results and comparison highlight the statistically significant performance improvement of our method in subspace representation, underscoring the benefits of utilizing NMF and supervised NMF subspaces as viable alternatives to SVD. 
Moreover, the part-based representation of the supervised NMF reveals the potential that SVD lacks in detecting discriminative information in medical imaging, as illustrated by saliency maps.

The rest of the paper is organized as follows. Section \ref{sec:Related-work} briefly reviews the related methods. Section \ref{Sec:Pipeline} details the few-shot learning framework on subspace feature representations, including the experimental settings and succinct descriptions of the datasets used. Extensive numerical experiments and comparison evaluating NMF and its variants in medical image classification, including detailed comparisons, are presented in Section \ref{sec:Experimental Results}. 
We conclude and point to some future work in Section \ref{sec:conclusion}.

\section{Related Work}\label{sec:Related-work}
In this section, we briefly review different subspace representations, including SVD, NMF, DNMF and SCNMFS, and a correspondence analysis method used for similarity comparison between different subspaces. 
In the following,  $\boldsymbol{X} = \left[\boldsymbol{x}_1, \boldsymbol{x}_2, \cdots, \boldsymbol{x}_N\right] \in \mathbb{R}^{M\times N}$ represents a data matrix, where $\boldsymbol{x}_i, 1\le i \le N$, are $M$-dimensional feature vectors, and $N$ is the number of feature vectors (or data samples).

\subsection{Singular Value Decomposition}\label{sec:SVD}
The SVD of the data matrix $\boldsymbol{X}$ can be written as
\begin{equation}
\boldsymbol{X}=\boldsymbol{U} \boldsymbol{\Sigma} \boldsymbol{P}^{\top},
\label{equation:SVD}
\end{equation}
where $\boldsymbol{U} = \left[\boldsymbol{u}_1, \cdots, \boldsymbol{u}_M\right] \in \mathbb{R}^{M \times M}$ and $\boldsymbol{P} \in \mathbb{R}^{N \times N}$ are unitary matrices, and $\boldsymbol{\Sigma} \in \mathbb{R}^{M \times N}$ is a diagonal matrix formed by $r$ singular values of $\boldsymbol{X}$; note that $r = {\rm rank}(\boldsymbol{X}) \leq \min\{M, N\}$. The columns of $\boldsymbol{U}$ and $\boldsymbol{P}$ are the so-called left and right singular vectors, respectively. 
Let $\boldsymbol{U}_k = \left[\boldsymbol{u}_1, \cdots, \boldsymbol{u}_k\right]$, $1 \leq k \leq r$. Utilising $\boldsymbol{U}_k$ as the projection matrix, the low-dimensional SVD subspace representation of $\boldsymbol{X}$ can be obtained by computing $\boldsymbol{V}=\boldsymbol{X}^{\top}\boldsymbol{U}_k$.

With the variance preserving property in compressing unimodal data, SVD has been used to interpret the learning dynamics of models across layers and models \cite{raghu2017svcca} and to find representative components in few-shot learning \cite{simon2020adaptive}.

\subsection{Non-negative Matrix Factorization} \label{NMF}

NMF is a well-known matrix factorization method that works by reconstructing a low-rank approximation of the input data matrix under the non-negativity constraint \cite{NIPS2000_f9d11525}. Given a non-negative data matrix $\boldsymbol{X} \in \mathbb{R}^{M\times N}$, the purpose of NMF is to seek two non-negative and low-rank matrices $\boldsymbol{U} \in \mathbb{R}^{M\times k}$ and $\boldsymbol{V} \in \mathbb{R}^{N \times k}$  under the condition of  $\boldsymbol{X} \approx \boldsymbol{U} \boldsymbol{V}^\top$, where $k < \min \{M, N\}$.
The non-negativity of all entries in $\boldsymbol{U}$ and $\boldsymbol{V}$ induces the sparsity of the subspace as well as the part-based representation. 
NMF can be formulated as the following constrained optimization problem
\begin{equation} 
\label{equation:NMF}
\min_{\boldsymbol{U, V}}\|\boldsymbol{X-U V}^\top\|_{F}^{2}, \quad
\text {s.t.} \ \boldsymbol{U} \geq 0, \boldsymbol{V} \geq 0,
\end{equation}
where $\|\cdot\|_{F^{}}$ is the Frobenius norm. The work in \cite{NIPS2000_f9d11525} proposed multiplicative iterative updating rules to find $\boldsymbol{U}$ and $\boldsymbol{V}$ for the minimization problem \eqref{equation:NMF}, i.e.,
\begin{equation}
\label{eqn:nmf-update}
u_{i j} \leftarrow u_{i j} \frac{(\boldsymbol{X} \boldsymbol{V})_{i j}}{\left(\boldsymbol{U V}^\top \boldsymbol{V}\right)_{i j}}, \quad
v_{i j} \leftarrow v_{i j} \frac{\left(\boldsymbol{X}^\top \boldsymbol{U}\right)_{i j}}{\left(\boldsymbol{V U}^\top \boldsymbol{U}\right)_{i j}}.
\end{equation} 
Note that $(\cdot)_{i j}$ represents the $(i,j)$ entry of the given matrix.

NMF has achieved tremendous success in various domains such as signal processing \cite{dong2021transferred}, biomedical engineering \cite{leuschner2019supervised}, pattern recognition \cite{lee1999learning}, and image processing \cite{chen2021deep}. It is unsupervised factorization without fully utilising the label information in classification tasks. Below we briefly recall some  supervised NMF methods, which will be investigated in our developed few-shot learning framework in Section \ref{Sec:Pipeline}.

\subsection{Discriminative Non-negative Matrix Factorization}\label{DNMF}

Babee et al. \cite{babaee2016discriminative} proposed the DNMF method, coupling discriminative regularizers generated from the labelled data with the main NMF objective function.
Using the discriminative constraint from labels, each class is dispersed into a separate cluster in the resulting subspace -- an appealing property for classification tasks.

For $C$ number of classes, the label matrix $\boldsymbol{Q} \in \mathbb{R} ^{ C \times N }$ is introduced with one-hot processing of the labels corresponding to the samples in $\boldsymbol{X}$. With an auxiliary matrix $\boldsymbol{A} \in \mathbb{R}^{C \times k}$, the aim of DNMF is to find $\boldsymbol{U}, \boldsymbol{V}$ and $\boldsymbol{A}$ satisfying
\begin{equation}
\label{equation:DNMF}
\begin{split}
    & \min_{\boldsymbol{U}, \boldsymbol{V}, \boldsymbol{A}}\|\boldsymbol{X}-\boldsymbol{U V}^\top\|_{F}^{2} + \alpha\|\boldsymbol{Q}-\boldsymbol{A V}^\top\|_{F}^{2}, \\
   & \quad {\rm s.t.} \ \boldsymbol{U} \geq \mathbf{0}, \boldsymbol{V} \geq \mathbf{0},
\end{split}    
\end{equation}
where $\alpha >0 $ is a constant used to balance the two terms in \eqref{equation:DNMF}. 
Note that $\boldsymbol{A}$ is allowed to take negative values. 
Analogous to the iterative formula in \eqref{eqn:nmf-update}, problem \eqref{equation:DNMF} can be solved by updating $\boldsymbol{U}, \boldsymbol{V}$ and $\boldsymbol{A}$ alternatively, as derived in \cite{babaee2016discriminative}, i.e.,
\begin{equation} \label{eqn:dnmf-update}
    \begin{gathered}
     u_{i j} \leftarrow u_{i j} \frac{(\boldsymbol{X V})_{i j}}{\left(\boldsymbol{U V}^\top \boldsymbol{V}\right)_{i j}} ,
     \\
     v_{i j} \leftarrow v_{i j} \frac{
     \left[\boldsymbol{X}^\top \boldsymbol{U} 
     + \alpha\left(\boldsymbol{V} \boldsymbol{A}^\top \boldsymbol{A}\right)^{-}
     +\alpha\left(\boldsymbol{Q}^\top \boldsymbol{A}\right)^{+}\right]_{i j}}
     {\left[\boldsymbol{V U}^\top \boldsymbol{U}
     +\alpha\left(\boldsymbol{V} \boldsymbol{A}^\top \mathbf{A}\right)^{+}
     +\alpha\left(\boldsymbol{Q}^\top \boldsymbol{A}\right)^{-}\right]_{i j}} ,
     \\
     \boldsymbol{A} \leftarrow \boldsymbol{Q} \boldsymbol{V}\left(\boldsymbol{V}^\top \boldsymbol{V}\right)^\dagger,
     \end{gathered}
\end{equation}
where the notations $(\cdot)^{+}$ and $(\cdot)^{-}$ represent the operations of setting the negative entries and positive entries in the given matrix to zero, respectively.

\subsection{Supervised and Constrained Non-negative Matrix Factorization with Sparseness}\label{SCNMFS}

The SCNMFS method proposed in \cite{cai2018supervised} is a variant of the constrained NMF \cite{liu2011constrained}. It takes both the sparsity and discriminative property into account under the condition of integrating the label information into the standard NMF decomposition. The insight into SCNMFS formulation resembles vector quantizing since it forces data with the same label to have the same latent representations \cite{liu2011constrained}. The hard constraints on labels ensure that the latent representations of the data samples from the same class are the same.

After observing the non-negative coefficient matrix $\boldsymbol{Z} \in \mathbb{R}^{C \times k}$ and the label matrix $\boldsymbol{Q}$, the SCNMFS subspace representation is expressed as $\boldsymbol{V}=\boldsymbol{Q}^\top \boldsymbol{Z}$. The reconstruction process is transferred into  $\boldsymbol{X} \approx \boldsymbol{U} \boldsymbol{V}^\top= \boldsymbol{U} \boldsymbol{Z}^\top \boldsymbol{Q}$. In our setting, the matrix  $\boldsymbol{U}$ is subjected to a Frobenius norm constraint, which is a commonly used approach in matrix optimization problems for regularizing the solution and avoiding overfitting. SCNMFS, then, is to find $\boldsymbol{U}$ and $\boldsymbol{Z}$ satisfying
\begin{equation}
\label{equation:SCNMFS}
   \min_{\boldsymbol{U,Z}}\|\boldsymbol{X-U Z}^\top \boldsymbol{Q}\|_{F}^{2}+\beta \|\boldsymbol{U}\|_{F}^{2}, \  \text{ s.t. } \boldsymbol{U} \geq 0, \boldsymbol{Z} \ \geq 0 , 
\end{equation}
where $\beta \in (0,1)$ balances the two terms in \eqref{equation:SCNMFS}. Analogous to the iterative formulas in \eqref{eqn:nmf-update} and \eqref{eqn:dnmf-update}, problem \eqref{equation:SCNMFS} can be solved by updating $\boldsymbol{U}$ and $\boldsymbol{Z}$ alternatively \cite{cai2018supervised}, i.e.,
\begin{equation}
\label{eqn:SCNMFS-update}
    \begin{gathered}
    u_{i j} \leftarrow u_{i j} \frac{(\boldsymbol{X Q}^\top \boldsymbol{Z})_{i j}}{\left(\boldsymbol{U Z}^\top \boldsymbol{Q Q}^\top \boldsymbol{Z}\right)_{i j}+\beta u_{i j}}, \\
    z_{i j} \leftarrow z_{i j} \frac{\left(\boldsymbol{Q} \boldsymbol{X}^{\top} \boldsymbol{U} \right)_{i j}}{\left(\boldsymbol{Q Q}^\top \boldsymbol{Z U}^{\top} \boldsymbol{U}\right)_{i j}} .
    \end{gathered}
\end{equation}
The final discriminative and sparse subspace is then generated by $\boldsymbol{V}=\boldsymbol{Q}^\top \boldsymbol{Z}$.
The learned projection matrix $\boldsymbol{U}$ with better sparsity ensures the identification capability of the obtained subspace $\boldsymbol{V}$.

\subsection{Canonical Correlation Analysis}
Canonical correlation analysis (CCA) \cite{hardoon2004canonical} is a multivariate method elucidating the correlation between two datasets by inferring information from a cross-covariance matrix. Given two data matrices $\boldsymbol{X_1} \in \mathbb{R}^{M_1 \times N}$ and $\boldsymbol{X_2} \in \mathbb{R}^{M_2 \times N}$, the whole process can be expressed as seeking vectors $\boldsymbol{a} \in \mathbb{R}^{M_{1}}$ and $\boldsymbol{b} \in \mathbb{R}^{M_{2}}$ maximising the correlation $\rho $, i.e.,
\begin{equation}
\label{equ:rho}
\begin{split}
\rho & =\operatorname {corr} (\boldsymbol{a}^{\top}\boldsymbol{X_1},\boldsymbol{b}^{\top}\boldsymbol{X_2}) \\
& = \frac{\boldsymbol{a}^{\top} \Sigma_{\boldsymbol{X_1} \boldsymbol{X_2}} \boldsymbol{b}}{\sqrt{\boldsymbol{a}^{\top} \Sigma_{\boldsymbol{X_1} \boldsymbol{X_1}} \boldsymbol{a}} \sqrt{\boldsymbol{b}^{\top} \Sigma_{\boldsymbol{X_2} \boldsymbol{X_2}} \boldsymbol{b}}},
\end{split}
\end{equation}
where $\Sigma_{\boldsymbol{X_1} \boldsymbol{X_1}}$, $\Sigma_{\boldsymbol{X_1} \boldsymbol{X_2}}$, and $\Sigma_{\boldsymbol{X_2} \boldsymbol{X_2}}$ are the cross-covariance matrices.

In this paper, we use CCA to compare the decomposition results of $\boldsymbol{U}$ and $\boldsymbol{V}$ obtained by different subspace representation methods.

\begin{figure}[tbp]
\centering
\includegraphics[width=3.4in,height=8.5cm]{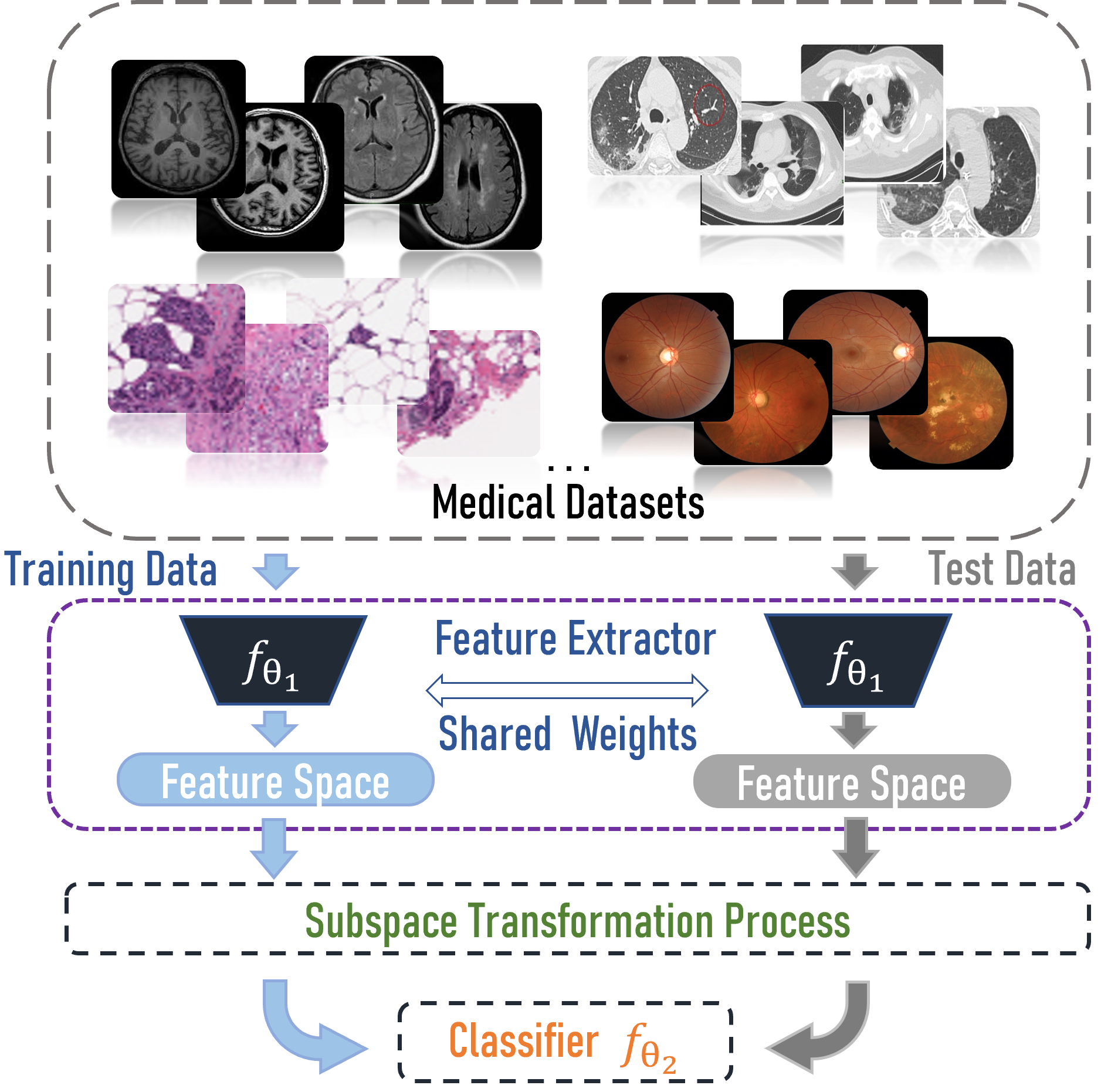}
\caption{Few-shot learning framework on medical imaging; see the main text for the detailed description. Note that there is no fine-tuning or training process for feature extraction (i.e., the purple block).}
\label{Diagram}
\end{figure}

\begin{algorithm}[htb] 
\caption{Few-shot learning framework} 
\label{alg:Framwork} 
\begin{algorithmic}[1] 
\REQUIRE $\boldsymbol{\cal S}_{\rm{train}}$,  $\boldsymbol{\cal S}_{\rm{test}}$, $
f_{\Theta_1}$, $k$ (i.e., subspace dimension), and Metype (i.e., method type) 
\ENSURE The predicted labels and accuracy of $\boldsymbol{\cal D}_{\rm{test}}$
\algrule
\STATE Generate feature spaces $\boldsymbol{X}_{\rm train}$ and $\boldsymbol{X}_{\rm test}$, i.e., $\boldsymbol{X}_{\rm train} \leftarrow { f_{\Theta_1}(\boldsymbol{D}_{\rm train})}$ and $\boldsymbol{X}_{\rm test} \leftarrow { f_{\Theta_1}(\boldsymbol{D}_{\rm test})}$;
\vspace{0.07in}

\STATE {Calculate} $\boldsymbol{U}_{\rm{train}}^{\Delta}$ and $\boldsymbol{V}_{\rm{train}}^{\Delta}$ in the \textbf{Switch} function below;
\SWITCH {MeType}
\CASELINE {SVD} using \eqref{equation:SVD};
\CASELINE {NMF} using \eqref{eqn:nmf-update};
\CASELINE {DNMF} using \eqref{eqn:dnmf-update};
\CASELINE {SCNMFS} using \eqref{eqn:SCNMFS-update};
\ENDSWITCH
\STATE Train the classifier $f_{\Theta_2}$ with $\boldsymbol{V}_{\rm{train}}^{\Delta}$ and $\boldsymbol{\cal Y}_{\rm{train}}$;
\vspace{0.07in}

\STATE Let $\boldsymbol{U}_{\rm{test}}^{\Delta} = \boldsymbol{U}_{\rm{train}}^{\Delta}$;\label{Step:9}
\SWITCH {MeType}
\CASELINE {SVD}  compute  $\boldsymbol{V}_{\rm{test}}^{\Delta}$ with $\boldsymbol{U}_{\rm{test}}^{\Delta}$ using \eqref{equation:SVD};
\CASELINE {NMF / DNMF / SCNMFS}
       \\ \qquad fix $\boldsymbol{U}_{\rm{test}}^{\Delta}$ and compute  $\boldsymbol{V}_{\rm{test}}^{\Delta}$ using \eqref{eqn:nmf-update};
\ENDSWITCH
\vspace{0.07in}

\STATE Compute the predicted labels of $\boldsymbol{\cal D}_{\rm{test}}$ using $f_{\Theta_2}$  with $\boldsymbol{V}_{\rm{test}}^{\Delta}$, and compute the accuracy using  $\boldsymbol{\cal Y}_{\rm{test}}$.
\end{algorithmic}
\end{algorithm}

\section{Method}\label{Sec:Pipeline}
In this section, we present the developed data-driven few-shot learning framework for medical image classification. Different from the existing few-shot learning methods, we exploit the function of NMF and its variants. Moreover, we also describe the data used in this research and the experiment settings. 

\begin{table*}[htbp] 
\centering
\caption{Details of the 14 medical datasets used in the experiments. }
\setlength{\tabcolsep}{-0.28mm}{
\begin{tabular}{p{25mm} p{100mm} c p{25mm} r}
\toprule
\multicolumn{1}{c}{Data}  &
  \multicolumn{1}{c}{Brief  introduction} &
  \# Classes &
  \multicolumn{1}{c}{Modality} &
  Scale \\ \midrule
{\tt BreastCancer}\cite{breastcancer} &
   Invasive ductal carcinoma (IDC), the most prevalent subtype of all breast cancers, is collected,  containing breast histopathology images.
  &
    2  &
    Histopathology &
  277,524 \\ \midrule
{\tt BrainTumor}\cite{cheng_2017}  &
  Brain tumor dataset provides human brain images including the intricate abnormalities in brain tumor size and location with 4 categories: glioma, meningioma, no tumor and pituitary.
   &
  4 &
  MRI &
  7,022 \\ \midrule
{\tt CovidCT}\cite{zhao2020COVID-CT-Dataset} &
 CovidCT contains hundreds of scans for COVID-19 with high diagnostic accuracy.
  &
  2 &
  CT &
  812 \\ \midrule
{\tt DeepDRiD} \cite{liu2022deepdrid} &
  Deep Diabetic Retinopathy dataset provides dual-view fundus images of eyes with discernible quality levels, containing information from 500 patients.
 &
  5 &
  Regular \newline refund  image &
  2,000 \\ \midrule
{\tt BloodMNIST}\cite{bloodmnist} &
  BloodMNIST dataset is built on a dataset of individual normal blood cells from 8 categories without infection during collection.
  &
  8 &
 Blood Cell \newline Microscope &
  17,092 \\ \midrule
{\tt BreastMNIST}\cite{breastmnist} &
BreastMNIST dataset is composed of breast ultrasound images. The original categorized classes (normal, benign, and malignant) are simplified into binary classification by combining normal and benign as positive  and classifying them against malignant as negative. We also explore the deep features class activation map with its provided segmentation masks.
 &
  2 &
 Breast  Ultrasound &
  780 \\ \midrule
{\tt DermaMNIST}\cite{tschandl2018ham10000,codella2018skin} &
  DermaMNIST dataset is regarding common pigmented skin lesions with $7$ distinct illnesses.
 &
  7 &
  Dermatoscope &
  10,015 \\ \midrule
{\tt OCTMNIST}\cite{kermany2018identifying} &
   OCTMNIST is based on valid optical coherence tomography (OCT) images for retinal diseases. &
  4 &
  Retinal OCT &
  109,309 \\ \midrule
\begin{tabular}[c]{@{}c@{}}{\tt OrganAMNIST}\cite{bilic2019liver} \\
{\tt OrganCMNIST}\cite{bilic2019liver} \\
{\tt OrganSMNIST}\cite{bilic2019liver}
\end{tabular} &
  \begin{tabular}[c]{@{}l@{}}OrganA/C/SMNIST datasets are generated from 3D CT images of Liver \\ Tumor Segmentation  Benchmark\cite{bilic2019liver}, based on cropping from the centre\\ slices  of 3D bounding boxes  in axial, coronal and sagittal views, respectively.
  \end{tabular} &
  11 &
  Abdominal  CT &
  \begin{tabular}[c]{@{}c@{}} 
  58,850\\ 23,660 \\ 25,221
 \end{tabular} 
   \\ \midrule
{\tt PathMNIST}\cite{kather2019predicting} &
  PathMNIST dataset is based on two previous studies (NCT-CRC-HE-100K and CRC-VAL-HE-7K) that used  colorectal cancer histology slides to predict survival.&
  9 &
  Colon \newline Pathology &
  107,180 \\ \midrule
{\tt PneumoniaMNIST}\cite{kermany2018identifying} &
  PneumoniaMNIST is a dataset of pediatric chest X-Ray images. &
  2 &
  Chest X-Ray &
  5,856 \\ \midrule
{\tt TissueMNIST}\cite{woloshuk2021situ} &
  TissueMNIST dataset is obtained from the Broad Bioimage Benchmark Collection \cite{woloshuk2021situ,ljosa2012annotated}  which focuses on the human kidney cortex cells, segmented from 3 reference tissue specimens and organized into 8 categories. &
  8 &
  Kidney Cortex \newline Microscope &
  236,386 \\ \bottomrule
\end{tabular}
}
\label{table:description}
\end{table*}

\subsection{Notation}\label{subsec:Notation}
Let $\boldsymbol{\cal S} = \{\boldsymbol{\cal D}, \boldsymbol{\cal Y}\}$ be a given image dataset, where $\boldsymbol{\cal D}$ and $\boldsymbol{\cal Y}$ denote the images and their class labels, respectively. They are then separated into training and test sets, i.e., $\boldsymbol{\cal S}_{\rm{train}} = \{\boldsymbol{\cal D}_{\rm{train}}, \boldsymbol{\cal Y}_{\rm{train}}\}$ and $\boldsymbol{\cal S}_{\rm{test}} = \{\boldsymbol{\cal D}_{\rm{test}}, \boldsymbol{\cal Y}_{\rm{test}}\}$. 
 Let $|\boldsymbol{\cal D}| = L_2$, i.e., the number of images in set $\boldsymbol{\cal D}$.
 Let $\boldsymbol{D} = (\boldsymbol{d}_1, \boldsymbol{d}_2, \cdots, \boldsymbol{d}_{L_2}) \in \mathbb{R}^{L_1\times L_2}$ be a matrix representing the whole images in $\boldsymbol{\cal D}$, where $\boldsymbol{d}_i \in \mathbb{R}^{L_1}$ represents an image by concatenating the image columns and $L_1$ is therefore the image size. 

Let $f_{\Theta_1}$ represent the pre-trained deep neural network which will be used to extract the prior information (i.e., features of our interest) from the given training/test images. For example, $\forall \boldsymbol{d}_i \in \boldsymbol{\cal D}$, $f_{\Theta_1}(\boldsymbol{d}_i)$ is the obtained feature vector of image $\boldsymbol{d}_i$. Then all the feature vectors obtained by $f_{\Theta_1}$ on set $\boldsymbol{\cal D}$ can form a feature space, being represented by matrix $\boldsymbol{X} = (\boldsymbol{x}_1, \boldsymbol{x}_2, \cdots, \boldsymbol{x}_{L_2}) = (f_{\Theta_1}(\boldsymbol{d}_1), f_{\Theta_1}(\boldsymbol{d}_2), \cdots, f_{\Theta_1}(\boldsymbol{d}_{L_2}))$. For simplicity, we define the process as $\boldsymbol{X} \leftarrow {f_{\Theta_1}(\boldsymbol{D})}$.
Analogously, the feature spaces corresponding to the training and test sets can be obtained by $\boldsymbol{X}_{\rm{train}} \leftarrow {f_{\Theta_1}(\boldsymbol{D}_{\rm{train}})}$ and $\boldsymbol{X}_{\rm{test}} \leftarrow {f_{\Theta_1}(\boldsymbol{D}_{\rm{test}})}$, respectively.

Let $\Delta$ represent the method SVD, NMF, DNMF or SCNMFS described in Section \ref{sec:Related-work}. 
Denote $\boldsymbol{U}_{\rm{train}}^{\Delta}$ and $\boldsymbol{V}_{\rm{train}}^{\Delta}$ (and $\boldsymbol{U}_{\rm{test}}^{\Delta}$ and $\boldsymbol{V}_{\rm{test}}^{\Delta}$) as the projection matrix and the subspace representation with $k$ columns, respectively, generated by method $\Delta$ at the training stage (and the test stage).
Let $ f_{\Theta_2}$ be the classifier trained on $\{\boldsymbol{V}_{\rm{train}}^{\Delta}, \boldsymbol{\cal Y}_{\rm{train}}\}$ and tested on $\{\boldsymbol{V}_{\rm{test}}^{\Delta}, \boldsymbol{\cal Y}_{\rm{test}}\}$.

\subsection{Framework}
Fig. \ref{Diagram} shows our proposed framework utilizing a pre-trained backbone network and an NMF dimensionality reduction scheme. The original medical dataset is randomly divided into a training set and a test set. The corresponding feature space is then extracted by the pre-trained network without fine-tuning steps. 
Following the dimensionality reduction of the feature space with the introduced different types of NMF methods, the connection between the subspaces of the training set and the test set is formed by implicit operations in the subspace transformation process.
Finally, one classifier is trained from the generated subspace and performs the prediction task.

To better understand the framework in Fig. \ref{Diagram}, the procedures are summarised in Algorithm \ref{alg:Framwork}. Firstly, the feature spaces corresponding to the training and test image sets are generated by $\boldsymbol{X}_{\rm{train}} \leftarrow {f_{\Theta_1}(\boldsymbol{D}_{\rm{train}})}$ and $\boldsymbol{X}_{\rm{test}} \leftarrow {f_{\Theta_1}(\boldsymbol{D}_{\rm{test}})}$, respectively. 
With the selected method $\Delta$ and the subspace dimension $k$, the projection matrix $\boldsymbol{U}_{\rm{train}}^{\Delta}$ and the subspace representation $\boldsymbol{V}_{\rm{train}}^{\Delta}$ are derived from the feature space $\boldsymbol{X}_{\rm{train}}$. 
The projection matrix $\boldsymbol{U}_{\rm{test}}^{\Delta}$ for the test data is yielded by the established implicit relationship between the training set and the test set (i.e., step \ref{Step:9} in Algorithm \ref{alg:Framwork}). 
The subspace representation $\boldsymbol{V}_{\rm{test}}^{\Delta}$ for the test data is then obtained by decomposing $\boldsymbol{X}_{\rm{test}}$ by  $\boldsymbol{U}_{\rm{test}}^{\Delta}$.
Classifier $ f_{\Theta_2}$ trained on the subspace representation $\boldsymbol{V}_{\rm{train}}^{\Delta}$ is finally applied on the subspace representation $\boldsymbol{V}_{\rm{test}}^{\Delta}$ to predict the final classification results.

\subsection{Data Description}\label{subsec:data}
A total of 14 different datasets covering a range of problems in diagnostics are employed for validation.  The detailed description of these datasets is given in Table \ref{table:description}.
The four selected medical datasets ({\tt BreastCancer}, {\tt BrainTumor}, {\tt CovidCT} and {\tt DeepDRiD}) cover diseases such as breast cancer \cite{breastcancer,cruz2014automatic}, brain tumor \cite{cheng_2017}, COVID-19 \cite{zhao2020COVID-CT-Dataset} and diabetic retinopathy \cite{liu2022deepdrid}. Early identification and classification of these diseases is an important research topic in medical imaging since it aids in the selection of the best treatment choices for patients. Ten datasets with {\tt MNIST} in their names are part of the {\tt MedMNIST} \cite{medmnistv2} benchmark collection. They cover primary data modalities (i.e., X-Ray, OCT, Ultrasound, CT and Electron Microscope) and diverse classification tasks (i.e., binary/multiclass with classes ranging from 2 to 11). 

Each image in our experiments is resized to $50 \times 50$ pixels and we randomly sample each of those datasets into two sizes (i.e., including $300$ and $600$ images) instead of using the entire datasets. The sizes of datasets are chosen to be on either side of the dimensionality of the feature space, which for ResNet is $512$. The greyscale images are converted into RGB images via the strategy in \cite{medmnistv2} to ensure the compatibility of the pre-trained network. It is noteworthy that the purpose of selecting these datasets is to illustrate the subspace mechanism of few-shot learning, rather than to compare with previous results on these datasets.

\begin{table*}[b]
\centering
\caption{Few-shot learning classification accuracy with KNN classifier. The original feature space and 30-dimensional subspaces obtained by SVD, NMF, DNMF and SCNMFS, including data augmentation, are evaluated on 14 medical datasets.}
\setlength{\tabcolsep}{1.3mm}{
\begin{threeparttable} 
\begin{tabular}{@{}ccccccccc@{}}
\toprule
&      & \multicolumn{6}{c}{Accurancy(\%)}   & \quad  \\ \midrule
Datasets  & Size & Augmentation & Feature  space & SVD & NMF & DNMF  & SCNMFS  & Classes  \quad           \\ \midrule
\multirow{2}{*}{{\tt BreastCancer}\cite{breastcancer}} 
& 300 & 40.75$\pm$6.25  & 65.00$\pm$2.37    & 70.25$\pm$4.43 & 62.00$\pm$4.37 & 69.00$\pm$6.63 & \textbf{77.75$\pm$3.48} & \multirow{2}{*}{2}  \\
& 600 & 42.75$\pm$4.30  & 69.00$\pm$2.87    & 72.25$\pm$1.51 & 65.25$\pm$2.64 & 70.38$\pm$2.19 & \textbf{75.32$\pm$2.12} &                     \\ \midrule
\multirow{2}{*}{{\tt BrainTumor}\cite{cheng_2017} } 
& 300  & 58.00$\pm$4.51 & 63.75$\pm$6.66 & 65.00$\pm$8.02 & 64.50$\pm$3.67 & 64.00$\pm$4.06 & \textbf{66.00$\pm$5.67} & \multirow{2}{*}{4}  \\
& 600  & 59.40$\pm$3.50 & \textbf{69.62$\pm$4.36} & 69.38$\pm$1.81 & 66.75$\pm$1.08 & 65.25$\pm$2.26 & 67.88$\pm$2.42 &                     \\ \midrule
\multirow{2}{*}{{\tt CovidCT}\cite{zhao2020COVID-CT-Dataset}}      
& 300  & 62.70$\pm$4.70 & \textbf{78.00$\pm$4.00} & 75.25$\pm$2.00 & 71.00$\pm$5.39 & 72.25$\pm$6.19 & 71.75$\pm$6.20          & \multirow{2}{*}{2}  \\
 & 600 & 62.12$\pm$3.72 & \textbf{81.50$\pm$1.46} & 79.25$\pm$2.81 & 77.50$\pm$1.53 & 76.12$\pm$2.28 & 70.75$\pm$1.74         &                     \\ \midrule
\multirow{2}{*}{{\tt DeepDRiD} \cite{liu2022deepdrid}}     
& 300  & 24.85$\pm$3.53 & 45.00$\pm$4.03          & 43.50$\pm$5.67 & 41.75$\pm$4.58 & 45.50$\pm$6.35 & \textbf{47.72$\pm$5.67} & \multirow{2}{*}{5}  \\
& 600  & 30.91$\pm$5.21 & \textbf{54.50$\pm$2.35} & 54.50$\pm$2.83 & 52.38$\pm$3.59 & 53.25$\pm$2.48 & 49.50$\pm$2.03          &                     \\ \midrule
\multirow{2}{*}{{\tt BloodMNIST}\cite{bloodmnist}}   
& 300  & 18.50$\pm$4.64 & 34.75$\pm$5.83          & 42.25$\pm$6.19 & 32.25$\pm$2.15 & 39.75$\pm$4.21 & \textbf{46.75$\pm$4.59} & \multirow{2}{*}{8}  \\
 & 600 & 15.12$\pm$2.45 & 37.62$\pm$2.78          & 45.00$\pm$3.10 & 39.75$\pm$2.64 & 41.50$\pm$2.70 & \textbf{48.88$\pm$2.78} &                     \\ \midrule
\multirow{2}{*}{{\tt BreastMNIST}\cite{breastmnist}}  
& 300  & 71.75$\pm$8.82 & \textbf{72.74$\pm$4.36} & 73.75$\pm$5.30 & 70.50$\pm$6.50 & 71.75$\pm$5.62 & 68.75$\pm$5.18          & \multirow{2}{*}{2}  \\
& 600  & 71.44$\pm$4.04 & \textbf{74.50$\pm$1.74} & 73.12$\pm$3.51 & 70.87$\pm$3.41 & 72.62$\pm$3.76 & 68.62$\pm$4.72          &                     \\ \midrule
\multirow{2}{*}{{\tt DermaMNIST}\cite{tschandl2018ham10000}}   
& 300  & 18.00$\pm$6.83 & 23.75$\pm$1.77          & 29.50$\pm$3.92 & 23.75$\pm$4.18 & 27.25$\pm$5.88 & \textbf{37.75$\pm$4.70} & \multirow{2}{*}{7}  \\
& 600  & 20.62$\pm$2.23 & 28.62$\pm$2.35          & 32.38$\pm$3.36 & 27.38$\pm$1.39 & 31.87$\pm$2.71 & \textbf{38.50$\pm$2.42} &                     \\ \midrule
\multirow{2}{*}{{\tt OCTMNIST}\cite{kermany2018identifying}}     
& 300  & 23.00$\pm$5.45 & 29.00$\pm$7.59          & 30.75$\pm$3.22 & 29.75$\pm$4.64 & 30.50$\pm$6.05 & \textbf{31.75$\pm$4.23} & \multirow{2}{*}{4}  \\
& 600  & 25.38$\pm$3.25 & 34.62$\pm$4.41          & 34.25$\pm$5.43 & 29.88$\pm$3.52 & 29.25$\pm$2.94 & \textbf{37.38$\pm$3.43} &                     \\ \midrule
\multirow{2}{*}{{\tt OrganAMNIST}\cite{bilic2019liver}}  
& 300  & 17.25$\pm$6.24 & 23.50$\pm$5.09          & 35.00$\pm$6.17 & 27.75$\pm$4.14 & 30.25$\pm$5.09 & \textbf{42.75$\pm$4.57} & \multirow{2}{*}{11} \\
& 600  & 17.12$\pm$4.50 & 34.62$\pm$2.58          & 40.88$\pm$2.64 & 34.63$\pm$3.15 & 39.12$\pm$4.36 & \textbf{46.12$\pm$1.50} &                     \\ \midrule
\multirow{2}{*}{{\tt OrganCMNIST}\cite{bilic2019liver}}  
& 300  & 20.00$\pm$1.11 & 21.50$\pm$5.50          & 29.50$\pm$4.91 & 20.25$\pm$1.66 & 28.00$\pm$1.27 & \textbf{35.75$\pm$3.59} & \multirow{2}{*}{11} \\
& 600  & 12.75$\pm$2.87 & 28.75$\pm$4.15          & 34.25$\pm$3.61 & 28.12$\pm$2.34 & 30.87$\pm$3.32 & \textbf{42.25$\pm$3.10} &                     \\ \midrule
\multirow{2}{*}{{\tt OrganSMNIST}\cite{bilic2019liver}}  
& 300  & 13.75$\pm$3.44 & 22.50$\pm$3.71          & 24.25$\pm$3.12 & 18.50$\pm$2.15 & 24.75$\pm$3.66 & \textbf{30.25$\pm$3.74} & \multirow{2}{*}{11} \\
& 600  & 13.88$\pm$1.33 & 25.25$\pm$2.26          & 28.75$\pm$1.63 & 26.25$\pm$2.90 & 27.00$\pm$2.22 & \textbf{32.38$\pm$1.00} &                     \\ \midrule
\multirow{2}{*}{{\tt PathMNIST}\cite{kather2019predicting}}    
& 300  & 18.75$\pm$6.02 & 28.75$\pm$4.33          & 45.75$\pm$4.44 & 34.75$\pm$6.24 & 44.00$\pm$6.19 & \textbf{54.75$\pm$6.04} & \multirow{2}{*}{9}  \\
& 600  & 17.00$\pm$2.94 & 33.25$\pm$2.18          & 45.13$\pm$2.54 & 39.62$\pm$6.37 & 44.50$\pm$1.74 & \textbf{54.62$\pm$1.79} &                     \\ \midrule
\multirow{2}{*}{{\tt PneumoniaMNIST}\cite{kermany2018identifying}} 
& 300  & 54.50$\pm$7.27 & 66.75$\pm$4.08 & 72.75$\pm$3.98 & 63.25$\pm$6.25 & 74.25$\pm$5.28 & \textbf{80.75$\pm$3.22} & \multirow{2}{*}{2} \\
 & 600 & 50.37$\pm$4.26 & 71.00$\pm$1.70          & 76.25$\pm$2.17 & 71.12$\pm$2.86 & 74.12$\pm$3.55 & \textbf{81.12$\pm$2.48} &                     \\ \midrule
\multirow{2}{*}{{\tt TissueMNIST}\cite{woloshuk2021situ}}  
& 300  & 10.63$\pm$15.30& 17.25$\pm$4.57          & 20.00$\pm$2.85 & 15.50$\pm$3.02 & 18.75$\pm$3.79 & \textbf{21.75$\pm$3.76} & \multirow{2}{*}{8}  \\
& 600  & 13.13$\pm$14.80 & 19.38$\pm$3.35          & 19.88$\pm$1.50 & 19.38$\pm$3.19 & 18.88$\pm$2.66 & \textbf{23.62$\pm$1.00} &                     \\ \bottomrule
\end{tabular}    
\end{threeparttable} 
}
\label{table:nmf-results}
\end{table*}

\begin{figure*}[th]
  \vspace{-0.2em}
  \centering
  \includegraphics[width=6.8in,height=4.8in]{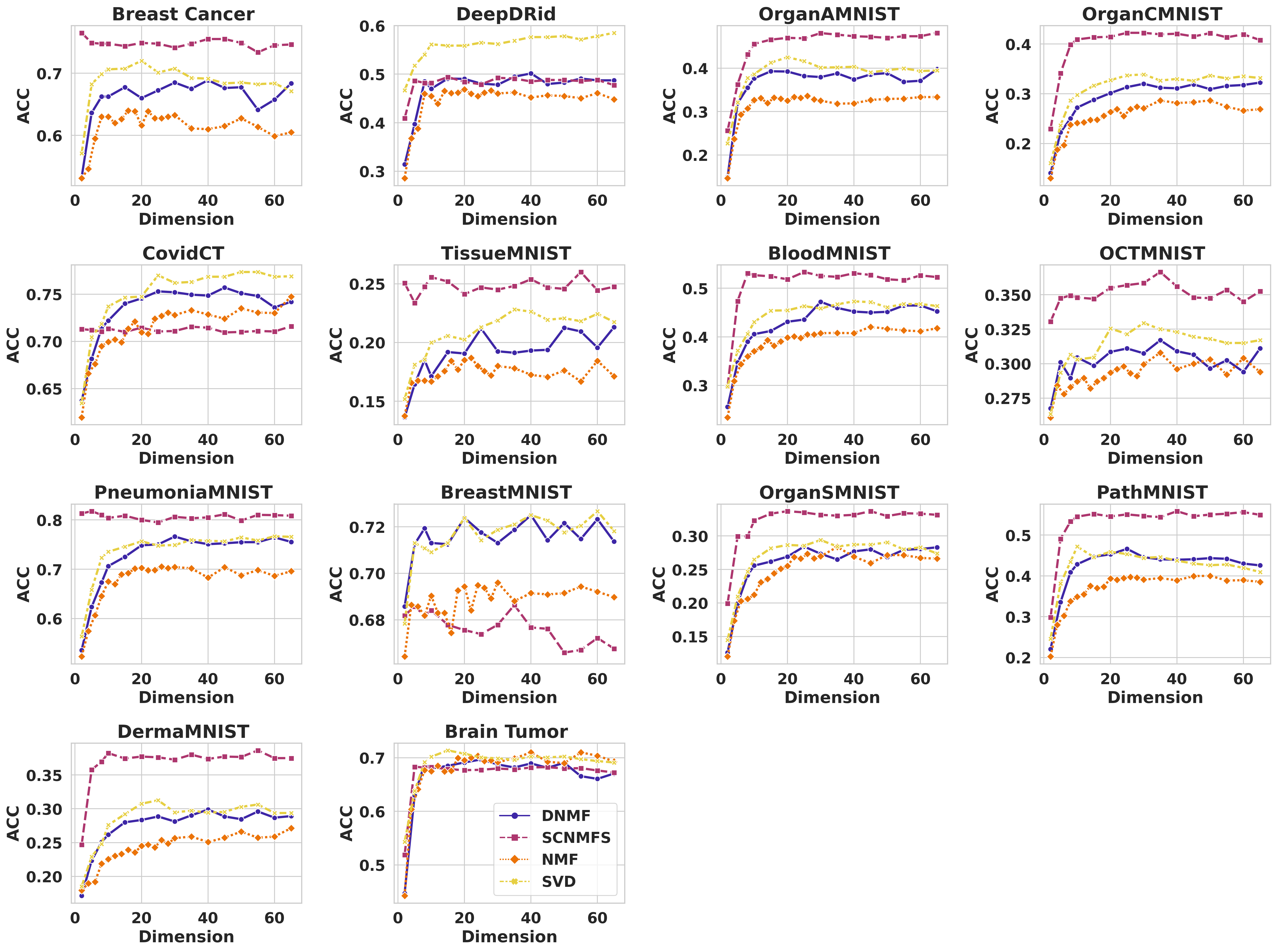}
  \caption{Classification accuracy of the NMF, DNMF, SCNMFS and SVD subspaces on 14 medical datasets with subspace dimensions ranging from 2 to 70. The dataset size is chosen as 600. In the plots, different colours correspond to different methods. Ten random partitions of the training-test set on each of the 14 datasets are conducted. It shows that the supervised NMF, especially SCNMFS, achieves significant improvements over the SVD-based subspace.}
  \label{Figure:all_results}
  \vspace{-0.4em}
\end{figure*}
\subsection{Experiment Setting}\label{subsec: Experiment Setting}

In our experiments, each dataset takes two sizes (i.e., 300 and 600 images) as the training sets, together with test sets with scales of 80 and 160 images, respectively. All the data are further preprocessed by subtracting the mean and being divided by the standard deviation before being used. Therefore, each image is distributed between 0 and 1 to eliminate the effect caused by abnormal data. ResNet18 pre-trained on ImageNet is used to generate the corresponding feature space. The feature space represents the output of the penultimate layer of ResNet18 implemented by PyTorch hooks \cite{pytorch}, yielding a $512$-dimensional feature vector for each image. 

The subspace representations of the extracted feature space are generated by the methods introduced in Section \ref{sec:Related-work}. 
The number of iterations related to NMF, DNMF and SCNMFS is set to 3000, ensuring the 
convergence. The main classifier used is the $K$-nearest neighbours (KNN) algorithm, where $K$ is set to $5$ unless otherwise specified. The reported average classification accuracy with standard deviation is achieved by repeating random samplings of the data 10 times.
Besides, we localise the distinct medical image regions (saliency map) from the subspace using the class activation map (CAM) method \cite{zhou2016learning}. The feature maps in the last convolutional layer are extracted using the same method as the feature space. To obtain visualizations maps, we take a fully connected layer as the classifier instead of KNN to find the weights and gradients. The detailed steps of implementing the CAM method are shown in Appendix \ref{appendix:1}. 

Comparisons are firstly conducted between the subspaces generated from SVD, NMF, DNMF and SCNMFS, with quantitative and qualitative metrics like classification accuracy and saliency maps. 
As widely believed that data augmentation techniques could bring positive impacts for the low data regime,
A comparison with data augmentation techniques \cite{AAC2023} and their impact on the feature space are also conducted in our experiments. Each image is subjected to standard rotation and cropping to increase the dataset size.
Ablation research involves comparing the feature space with different subspaces through dimensionality reduction, as conducted in Section \ref{subsec:effect of subspace}.
Moreover, we also compare our method with a well-known few-shot learning method -- prototypical network  \cite{snell2017prototypical}. The network is composed of four convolution blocks.
Each block is composed of a 64-filter $3 \times 3$ convolution, batch normalization layer, a ReLU nonlinearity and a $2 \times 2$ max-pooling layer. 
In our experiments, we pre-trained the network on the omniglot dataset \cite{lake2019omniglot} via SGD with Adam \cite{kingma2014adam} and obtained 99\% accuracy in the 10-shot scenario.

In the experiments, the ``$C$-way $l$-shot" setting means $l$ samples are given for each class in the support set, and $l$ query images per class are provided to validate the final performance.
The original feature space represents the features extracted by the pre-trained network without dimension reduction.  Different number of dimensions for the SVD/NMF/DNMF/SCNMFS subspaces are tested.  The final classification accuracy is computed by averaging over 10 randomly generated episodes from each medical dataset.

\begin{table*}[]\scriptsize
%
\setlength{\tabcolsep}{0.75mm}{
\centering

\caption{ Comparison between the prototypical network and the few-shot learning with subspace feature representations. Note that $C$ is the number of classes in each dataset and Dim stands for the subspace dimensions. }
\label{fsw-all}
\resizebox{2.15\columnwidth}{10.2cm}{
\begin{tabular}{ccccccccccc} 
\hline
\multicolumn{11}{c}{$C$-way 10-shot  Accuracy (\%)} \\ 
\hline
\multirow{3}{*}{Data}& \multicolumn{10}{c}{Methods}  \\ 
\cline{2-11}
& \multirow{2}{*}{\begin{tabular}[c]{@{}c@{}}Prototypical\\ Network\end{tabular}} & \multicolumn{9}{c}{Few-shot Learning with Subspace Feature Representations (\textbf{Ours})}\\ 
\cline{3-11}
& & Feature Space & Subspaces & 2 Dim   & 5 Dim   & 10 Dim 
&20 Dim    & 30 Dim   & 40 Dim   & 50 Dim   \\ 
\hline
\multirow{4}{*}{\begin{tabular}[c]{@{}c@{}}CovidCT\\ ($C$=2)\end{tabular}}        
& \multirow{4}{*}{51.00$\pm$9.78} & \multirow{4}{*}{51.00$\pm$8.17}   
& SVD   & 52.67$\pm$7.72   & 52.33$\pm$6.51   & 51.33$\pm$8.97   & 51.33$\pm$8.97  & 51.33$\pm$8.97  & 51.33$\pm$8.97  & 51.33$\pm$8.97           \\
& &
& NMF   & 49.67$\pm$8.23   & 56.67$\pm$7.45   & 56.67$\pm$8.56   & 58.00$\pm$8.06  & 57.00$\pm$5.68 & 55.33$\pm$6.70  & \textbf{58.00$\pm$7.18}  \\
& &
& DNMF& 48.33$\pm$7.92   & 53.00$\pm$5.47     & 53.00$\pm$8.23   & 49.00$\pm$6.84  & 51.33$\pm$11.18 & 50.33$\pm1$0.16 & 48.00$\pm$9.21            \\
& &
& SCNMFS & 49.00$\pm$9.32  & 51.00$\pm$9.78   & 51.33$\pm$8.33   & 52.70$\pm$9.04  & 51.33$\pm$8.33 & 51.67$\pm$8.47   & 52.70$\pm$9.17           \\ 
\hline
\multirow{4}{*}{\begin{tabular}[c]{@{}c@{}}BreastCancer\\ ($C$=2)\end{tabular}}   
& \multirow{4}{*}{70.00$\pm$12.45}   & \multirow{4}{*}{74.00$\pm$11.14}  
& SVD       & 66.00$\pm$15.94  & 71.5$\pm$11.84  & 72.50$\pm$10.31 & 74.50$\pm$9.86  & 74.50$\pm$9.86       & 74.50$\pm$9.86   & 74.50$\pm$9.86   \\
& &      
& NMF   & 66.00$\pm$14.46  & 74.50$\pm$9.60   & \textbf{76.00$\pm$8.60}  & 77.00$\pm$7.14  & 77.00$\pm$6.40       & 74.50$\pm$7.23   & 74.50$\pm$7.23   \\
&       &      & DNMF      & 66.00$\pm$11.14  & 74.50$\pm$9.07   & 70.00$\pm$12.25 & 72.00$\pm$9.27  & 71.00$\pm$7.35       & 69.50$\pm$10.36  & 69.00$\pm$8.89   \\
&       &      & SCNMFS    & 69.50$\pm$12.13  & 70.50$\pm$12.34  & 70.50$\pm$12.34 & 71.00$\pm$12.41 & 71.00$\pm$12.41       & 70.50$\pm$12.34  & 71.00$\pm$12.61  \\ 
\hline
\multirow{4}{*}{\begin{tabular}[c]{@{}c@{}}PneumoniaMNIST\\ ($C$=2)\end{tabular}} & \multirow{4}{*}{70.0$\pm$10.25}   & \multirow{4}{*}{73.5$\pm$7.09}   & SVD       & 65.00$\pm$10.72  & 73.00$\pm$8.12  & 75.00$\pm$7.75 & 74.00$\pm$7.68 & 74.00$\pm$7.68      & 74.00$\pm$7.68 & 74.00$\pm$7.68 \\
&       &      & NMF       & 61.00$\pm$11.14  & 82.50$\pm$6.80  & 82.50$\pm$6.42 & 82.00$\pm$7.14 & 82.50$\pm$5.59      & \textbf{84.00$\pm$4.90} & 83.00$\pm$6.00 \\
&       &      & DNMF      & 62.00$\pm$12.69  & 77.50$\pm$9.01  & 75.00$\pm$5.00 & 72.50$\pm$8.73 & 77.00$\pm$6.00      & 79.00$\pm$4.90 & 73.50$\pm$8.38 \\
&       &      & SCNMFS    & 70.00$\pm$11.4  & 71.50$\pm$10.50 & 75.00$\pm$8.37 & 73.50$\pm$9.76 & 73.50$\pm$9.50      & 73.50$\pm$10.50& 74.00$\pm$10.91\\ 
\hline
\multirow{4}{*}{\begin{tabular}[c]{@{}c@{}}BreastMNIST\\ ($C$=2)\end{tabular}}    & \multirow{4}{*}{59.5$\pm$13.68}   & \multirow{4}{*}{62.5$\pm$10.31}  & SVD       & 53.50$\pm$11.19 & 58.00$\pm$10.30 & 60.00$\pm$14.49& 62.00$\pm$9.54 & 62.00$\pm$9.54      & 62.00$\pm$9.54 & 62.00$\pm$9.54 \\
&       &      & NMF       & 56.00$\pm$11.58 & 66.50$\pm$9.76  & 69.00$\pm$10.91& \textbf{69.00$\pm$9.95} & 65.00$\pm$13.23     & 68.00$\pm$12.88& 67.00$\pm$14.87\\
&       &      & DNMF      & 56.00$\pm$10.44 & 60.00$\pm$12.04 & 56.50$\pm$16.44& 61.00$\pm$14.11& 61.50$\pm$12.66     & 62.50$\pm$11.67& 59.50$\pm$8.80  \\
&       &      & SCNMFS    & 62.00$\pm$10.30 & 62.50$\pm$9.81  & 64.50$\pm$9.34 & 64.50$\pm$7.57 & 64.50$\pm$8.20       & 63.50$\pm$10.26& 66.00$\pm$7.00  \\ 
\hline
\multirow{4}{*}{\begin{tabular}[c]{@{}c@{}}DeepDRid\\ ($C$=5)\end{tabular}}       & \multirow{4}{*}{33.00$\pm$6.02}    & \multirow{4}{*}{29.00$\pm$5.95}   & SVD       & 28.40$\pm$6.05   & 29.60$\pm$4.45   & 29.00$\pm$5.31  & 28.40$\pm$6.97  & 28.60$\pm$6.26       & 28.80$\pm$6.00   & 29.00$\pm$5.95  \\
&       &      & NMF       & 30.40$\pm$6.44   & 32.40$\pm$3.98   & 32.40$\pm$3.98  & 33.20$\pm$4.21  & 34.00$\pm$4.38       & 34.20$\pm$5.90  & 34.80$\pm$5.60  \\
&       &      & DNMF      & 28.40$\pm$6.05   & 29.20$\pm$4.40   & 29.00$\pm$2.72  & 30.00$\pm$6.07  & 29.40$\pm$4.90 & 30.40$\pm$5.64  & 30.80$\pm$5.00  \\
&       &      & SCNMFS    & 34.00$\pm$5.29   & 37.00$\pm$6.08   & 36.60$\pm$5.59  & 38.00$\pm$5.87  & \begin{tabular}[c]{@{}c@{}}\textbf{38.40$\pm$7.36}\\\end{tabular} & 37.00$\pm$6.08  & 37.00$\pm$5.16   \\ 
\hline
\multirow{4}{*}{\begin{tabular}[c]{@{}c@{}}BrainTumor\\ ($C$=4)\end{tabular}}     & \multirow{4}{*}{59.75$ \pm$ 4.25}  & \multirow{4}{*}{56.75 $\pm$ 8.95} & SVD       & 52.50$\pm$10.19  & 56.00$\pm$10.26  & 57.75$\pm$9.78  & 56.75$\pm$10.25 & 56.75$\pm$9.75       & 56.50$\pm$9.37  & 56.50$\pm$9.37  \\
&       &      & NMF       & 52.75$\pm$8.02   & 60.00$\pm$8.94   & 64.00$\pm$7.76  & 61.25$\pm$4.37  & 62.25$\pm$6.37       & 61.25$\pm$6.91  & 61.75$\pm$4.75  \\
&       &      & DNMF      & 51.50$\pm$9.23   & 56.75$\pm$9.22   & 59.25$\pm$7.08  & 58.50$\pm$6.34  & 58.25$\pm$8.22       & 56.75$\pm$6.13  & 58.00$\pm$7.05  \\
&       &      & SCNMFS    & 44.75$\pm$5.75   & 62.50$\pm$4.61   & 62.00$\pm$5.57  & 62.00$\pm$5.22  & \textbf{62.75$\pm$5.53}     & 62.20$\pm$5.53   & 62.75$\pm$6.08  \\ 
\hline
\multirow{4}{*}{\begin{tabular}[c]{@{}c@{}}BloodMNIST\\ ($C$=8)\end{tabular}}     & \multirow{4}{*}{4.88$\pm$5.38}    & \multirow{4}{*}{55.5$\pm$6.94}   & SVD       & 40.00$\pm$3.58  & 51.12$\pm$6.62  & 50.62$\pm$5.65~& 53.62$\pm$6.43 & 55.12$\pm$6.97      & 55.38$\pm$6.02 & 55.62$\pm$6.99 \\
&       &      & NMF       & 40.25$\pm$5.53  & 51.88$\pm$6.36  & 53.50$\pm$5.09 & 56.50$\pm$5.83 & 57.25$\pm$4.14      & 55.50$\pm$5.31 & 56.75$\pm$5.07 \\
&       &      & DNMF      & 39.50$\pm$6.30  & 48.87$\pm$6.41  & 51.12$\pm$5.08 & 51.87$\pm$5.51 & 49.50$\pm$5.48      & 52.62$\pm$6.34 & 52.12$\pm$5.81 \\
&       &      & SCNMFS    & 39.38$\pm$6.13  & 54.38$\pm$6.60  & 56.12$\pm$4.89 & 56.12$\pm$5.17 & \textbf{56.75$\pm$6.35}    & 55.75$\pm$5.71 & 54.50$\pm$5.71  \\ 
\hline
\multirow{4}{*}{\begin{tabular}[c]{@{}c@{}}DermaMNIST\\ ($C$=7)\end{tabular}}     & \multirow{4}{*}{29.14$\pm$5.83}   & \multirow{4}{*}{33.00$\pm$6.31}   & SVD       & 21.86$\pm$5.61  & 28.57$\pm$6.23  & 31.14$\pm$6.38 & 33.00$\pm$7.65  & 33.57$\pm$6.00       & 33.29$\pm$5.75 & 33.14$\pm$6.32 \\
&       &      & NMF       & 24.43$\pm$5.36  & 31.43$\pm$4.69  & 35.43$\pm$5.78 & 34.14$\pm$5.48 & 35.86$\pm$6.04      & 35.29$\pm$5.42 & 35.57$\pm$7.07 \\
&       &      & DNMF      & 23.00$\pm$5.17   & 30.71$\pm$6.55  & 32.14$\pm$4.88 & 34.00$\pm$5.14  & 31.00$\pm$8.31       & 32.14$\pm$6.52 & 29.29$\pm$8.21 \\
&       &      & SCNMFS    & 23.57$\pm$5.93  & 33.29$\pm$4.70   & \textbf{35.86$\pm$5.05} & 35.14$\pm$5.20  & 35.57$\pm$3.86      & 35.14$\pm$4.15 & 35.57$\pm$4.88 \\ 
\hline
\multirow{4}{*}{\begin{tabular}[c]{@{}c@{}}OCTMNIST\\ ($C$=4)\end{tabular}}       & \multirow{4}{*}{25.0$\pm$9.22}    & \multirow{4}{*}{28.25$\pm$6.32}  & SVD       & 27.50$\pm$5.00  & 25.25$\pm$6.84  & 26.75$\pm$5.25 & 27.75$\pm$6.84 & 28.00$\pm$5.6       & 28.50$\pm$6.34 & 28.50$\pm$6.34 \\
&       &      & NMF       & 28.25$\pm$6.90  & 31.50$\pm$6.04  & 33.00$\pm$7.81 & 34.00$\pm$5.15 & \textbf{36.50$\pm$3.74}    & 36.25$\pm$6.64 & 35.25$\pm$4.67 \\
&       &      & DNMF      & 26.50$\pm$7.67  & 26.50$\pm$7.84  & 27.25$\pm$7.78 & 33.00$\pm$6.87 & 31.25$\pm$7.35      & 29.25$\pm$6.33 & 30.75$\pm$4.88 \\
&       &      & SCNMFS    & 26.00$\pm$5.94  & 28.75$\pm$7.35  & 29.25$\pm$8.44 & 28.75$\pm$8.46 & 30.00$\pm$7.83      & 28.50$\pm$7.76 & 29.50$\pm$6.50 \\ 
\hline
\multirow{4}{*}{\begin{tabular}[c]{@{}c@{}}OrganAMNIST\\ ($C$=11)\end{tabular}}   & \multirow{4}{*}{55.18$\pm$3.57}   & \multirow{4}{*}{63.73$\pm$3.91}  & SVD       & 36.64$\pm$3.81  & 52.55$\pm$4.02  & 61.18$\pm$3.90 & 62.82$\pm$3.76 & 63.27$\pm$4.15      & 63.00$\pm$3.55 & 63.36$\pm$4.01 \\
&       &      & NMF       & 36.36$\pm$2.73  & 55.36$\pm$4.31  & 62.27$\pm$4.01 & 65.27$\pm$3.94 & 65.18$\pm$5.18      & \textbf{65.18$\pm$4.21} & 65.18$\pm$4.76 \\
&       &      & DNMF      & 35.18$\pm$3.20  & 51.73$\pm$5.98  & 61.09$\pm$3.94 & 63.09$\pm$3.62 & 64.27$\pm$5.32      & 61.82$\pm$5.01 & 59.64$\pm$4.40 \\
&       &      & SCNMFS    & 30.18$\pm$3.44  & 48.91$\pm$2.93  & 58.91$\pm$3.14 & 61.00$\pm$2.92 & 61.45$\pm$1.92      & 61.27$\pm$2.31 & 60.64$\pm$2.15 \\ 
\hline
\multirow{4}{*}{\begin{tabular}[c]{@{}c@{}}OrganCMNIST\\ ($C$=11)\end{tabular}}   & \multirow{4}{*}{49.64$\pm$3.28}   & \multirow{4}{*}{60.18$\pm$6.27}  & SVD       & 27.82$\pm$2.88  & 43.82$\pm$4.67  & 53.27$\pm$4.49 & 58.73$\pm$6.09 & 59.55$\pm$6.05      & 59.55$\pm$6.20 & 60.18$\pm$6.64 \\
&       &      & NMF       & 28.09$\pm$3.51  & 43.73$\pm$5.55  & 56.64$\pm$2.76 & 61.36$\pm$4.25 & \textbf{63.18$\pm$4.69}    & 62.18$\pm$4.34 & 62.91$\pm$4.45 \\
&       &      & DNMF      & 27.27$\pm$3.98  & 41.27$\pm$5.84  & 53.91$\pm$4.26 & 60.55$\pm$4.38 & 58.64$\pm$4.92      & 57.82$\pm$4.29 & 54.09$\pm$5.23 \\
&       &      & SCNMFS    & 27.18$\pm$3.89  & 43.82$\pm$3.72  & 53.82$\pm$4.06 & 56.36$\pm$4.21 & 56.09$\pm$4.05      & 56.73$\pm$3.62 & 55.73$\pm$4.49 \\ 
\hline
\multirow{4}{*}{\begin{tabular}[c]{@{}c@{}}OrganSMNIST\\ ($C$=11)\end{tabular}}   & \multirow{4}{*}{38.0$\pm$4.51}    & \multirow{4}{*}{43.82$\pm$4.37}  & SVD       & 24.45$\pm$2.92  & 38.36$\pm$6.73  & 41.18$\pm$5.46 & 42.09$\pm$4.64 & 42.91$\pm$4.39      & 43.91$\pm$4.44 & 44.00$\pm$4.21  \\
&       &      & NMF       & 26.00$\pm$3.73  & 40.36$\pm$3.82  & 42.27$\pm$3.69~& 45.73$\pm$3.02 & 45.18$\pm$2.41      & \textbf{46.00$\pm$2.82}  & 45.73$\pm$3.84 \\
&       &      & DNMF      & 25.00$\pm$3.28  & 37.64$\pm$4.31  & 40.36$\pm$3.75 & 40.82$\pm$4.90 & 41.27$\pm$5.91      & 41.36$\pm$4.44 & 42.55$\pm$3.06 \\
&       &      & SCNMFS    & 24.27$\pm$3.07  & 37.91$\pm$4.86  & 41.09$\pm$4.74 & 41.91$\pm$4.70 & 42.18$\pm$4.71      & 42.00$\pm$3.92  & 43.45$\pm$4.47 \\ 
\hline
\multirow{4}{*}{\begin{tabular}[c]{@{}c@{}}PathMINST\\ ($C$=9)\end{tabular}}      & \multirow{4}{*}{36.78$\pm$4.02}   & \multirow{4}{*}{42.56$\pm$3.88}  & SVD       & 26.11$\pm$2.91  & 36.67$\pm$3.65  & 40.44$\pm$3.52 & 42.67$\pm$4.16 & 41.89$\pm$4.87      & 42.44$\pm$4.06 & 42.56$\pm$4.04 \\
&       &      & NMF       & 28.00$\pm$4.89  & 39.89$\pm$3.28  & 44.22$\pm$3.74 & 45.78$\pm$2.85 & 45.56$\pm$2.58      & 43.56$\pm$4.06 & 43.56$\pm$4.33 \\
&       &      & DNMF      & 27.33$\pm$4.51  & 35.00$\pm$4.01  & 40.67$\pm$4.13 & 41.33$\pm$4.73 & 42.11$\pm$4.70      & 39.56$\pm$4.59 & 39.11$\pm$4.63 \\
&       &      & SCNMFS    & 26.22$\pm$3.19  & 40.56$\pm$3.11  & 41.89$\pm$3.52 & \textbf{43.89$\pm$3.58} & 42.56$\pm$4.28      & 43.00$\pm$3.26 & 41.22$\pm$4.37 \\ 
\hline
\multirow{4}{*}{\begin{tabular}[c]{@{}c@{}}TissueMNIST\\ ($C$=8)\end{tabular}}    & \multirow{4}{*}{25.5$\pm$3.67}    & \multirow{4}{*}{24.5$\pm$3.63}   & SVD       & 20.50$\pm$2.32  & 22.50$\pm$3.58  & 22.37$\pm$3.86 & 24.12$\pm$3.92 & 23.75$\pm$3.01      & 23.88$\pm$3.97 & 24.00$\pm$3.98 \\
&       &      & NMF       & 20.88$\pm$4.03  & 26.88$\pm$4.55  & 26.25$\pm$3.62 & 25.25$\pm$3.82 & 26.88$\pm$3.17      & 26.00$\pm$2.29 & 24.62$\pm$3.58 \\
&       &      & DNMF      & 19.25$\pm$3.84  & 22.62$\pm$4.73  & 23.12$\pm$3.55 & 23.75$\pm$2.37 & 22.25$\pm$4.96      & 21.25$\pm$4.71 & 22.12$\pm$3.83 \\
&       &      & SCNMFS    & 23.38$\pm$4.58  & 28.00$\pm$3.54  & 28.50$\pm$3.44 & \textbf{28.75$\pm$3.31} & 27.50$\pm$3.16      & 27.88$\pm$3.92 & 27.75$\pm$3.30 \\
\hline
\end{tabular}
}
}
\end{table*}
\section{Experimental Results and Discussion}\label{sec:Experimental Results} 
In this section, we evaluate the effectiveness of using subspaces in data-based few-shot learning for medical images with limited samples, and evaluate the part-based representation of NMF and its variants in detecting the distinct lesion area in medical imaging.

\subsection{Classification Results and Discussion}\label{subsec:Classification accuracy analysis}
\subsubsection{Effect of subspace}\label{subsec:effect of subspace}
 Table \ref{table:nmf-results} shows the classification results of using the original feature space and four subspaces obtained by SVD, NMF, DNMF and SCNMFS, including data augmentation. Both the orignal feature space and the feature space with data augmentation have 512 dimensions and are derived directly from the pre-trained network.
 The dimension of the subspaces derived by SVD, NMF, DNMF and SCNMFS is kept at $30$. As indicated before, experiments are conducted at two different training set sizes of $300$ and $600$ to explore the cases that the data size is smaller and larger than the dimension of the feature space (i.e., $512$).
In general, an increase in the amount of data affects the classification results, as more various features are brought in. Interestingly, the results in Table \ref{table:nmf-results} show that the subspace does not appear to have the same trend of improvement as the original feature space. The subspace representations (i.e., using SVD, NMF, DNMF and SCNMFS) overall yield better results than the original feature space (except for the datasets {\tt CovidCT} and {\tt BrainTumor}), indicating that dimensionality reduction is an important step for working in low data regimes. 
Table \ref{table:nmf-results} also shows that
data augmentation does not increase but decrease the classification performance, indicating its negative impact on the classification tasks here in medical imaging.

\begin{figure}[!htp]
\subfigure[ROC curve in 5D subspace.]{
\centering
\label{fig:ROC:5}
\includegraphics[width=0.48\linewidth]{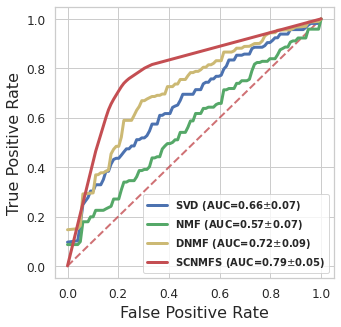}
}%
\subfigure[ROC curve in 15D subspace.]{
\centering
\label{fig:ROC:15}
\includegraphics[width=0.48\linewidth]{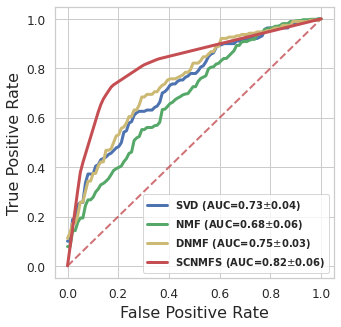}
}%
\caption{ROC curve of different subspaces (i.e., SVD, NMF, DNMF and SCNMFS) in 5D and 15D subspace representations on the {\tt PneumoniaMNIST} dataset with the size of 300. The blue, green, yellow and red lines represent the KNN results on SVD, NMF, DNMF and SCNMFS subspaces, respectively.  It shows that the performance of SCNMFS is much more stable than others including SVD in both low and high dimensions (i.e., 5D and 15D).} 
\label{fig:ROC-results}
\end{figure}

As shown in Fig. \ref{Figure:all_results}, we see that using subspaces is robust with different dimensions starting from 10. 
Table \ref{fsw-all} show that subspaces with appropriate dimensions are undoubtedly a better way with even only 10 samples per class available. Our proposed method also outperforms the prototypical network in all the datasets.

\subsubsection{Performance of SVD and NMF}
Regarding the performance of subspaces in few-shot learning on both the binary class and multiclass problems, the classification result discrepancies between the standard NMF and SVD are modest as shown in Table \ref{table:nmf-results}, with SVD performing slightly better. The selected rank during decomposition has an impact on the efficacy of the subspace when NMF converts data into sparse and part-based subspaces. As shown in Fig. \ref{Figure:all_results}, NMF yields stable and reliable results when the subspace dimension is not too small (e.g. $> 10$). 
SVD's performance is also limited by the number of dimensions, with features from its lower dimensions yielding poor performance in classification tasks.

The ROC results from the KNN classifier in multiple subspaces are shown in Fig. \ref{fig:ROC-results}. In this experiment, the {\tt PneumoniaMNIST} dataset of size $300$ is used, and two different dimensions (i.e., $5$ and $15$) are reported as an example. Different colour lines represent the average results of the different subspace methods. It shows that the performance of the SVD is better than the NMF subspaces (see the blue and green lines in Fig. \ref{fig:ROC-results}).
Interestingly, in Table \ref{fsw-all}, NMF shows its advantage compared to SVD when fewer data is available (e.g., 10 images per class) on the 14 medical datasets. 
This may indicate that the subspace obtained by SVD may not be as meaningful as NMF until the data size reaches a certain scale. Under appropriate dimensions, NMF can perform better than SVD. The sparse and part-based representation obtained by NMF can effectively preserve the original information in the subspace of the few-shot learning mechanism. Therefore, standard NMF can be a viable alternative to SVD when sparse subspace representation is of great interest.

\subsubsection{Performance of the supervised NMF}
The results of using the subspaces obtained by the supervised NMF methods (i.e., DNMF and SCNMFS) in few-shot learning are also given in Tables \ref{table:nmf-results} and \ref{fsw-all}. 
In Table \ref{table:nmf-results}, compared with NMF, the performance of the supervised NMF is significantly improved, indicating the effectiveness of incorporating the label information into the decomposition process.
Moreover, it also shows that the subspace generated from DNMF exhibits a slight advantage over the standard NMF and the performance of SCNMFS is more robust compared with the original feature space and other subspaces. This is also in line with the results of DNMF (yellow line) and SCNMFS (red line) shown in Fig \ref{fig:ROC-results}. Both DNMF and SCNMFS perform better than NMF in different subspace dimensions. Compared to SVD, the effect of SCNMFS is more stable. 
In Table \ref{fsw-all}, the performance of the supervised NMF method is not that impressive compared to the standard NMF. This tells us that incorporating label signals into the decomposition process of NMF will be affected by the data size. Considering the real case where the number of images in most medical datasets are hundreds available, the subspace generated by the supervised NMF method is quite a competitive solution.

\begin{figure}[!th]
\vspace{0.6em}
\subfigure[Projection matrix $\boldsymbol{U}_{\rm train}^{\Delta}$.]{
\centering
\label{fig:ROC:5a}
\includegraphics[width=0.48\linewidth]{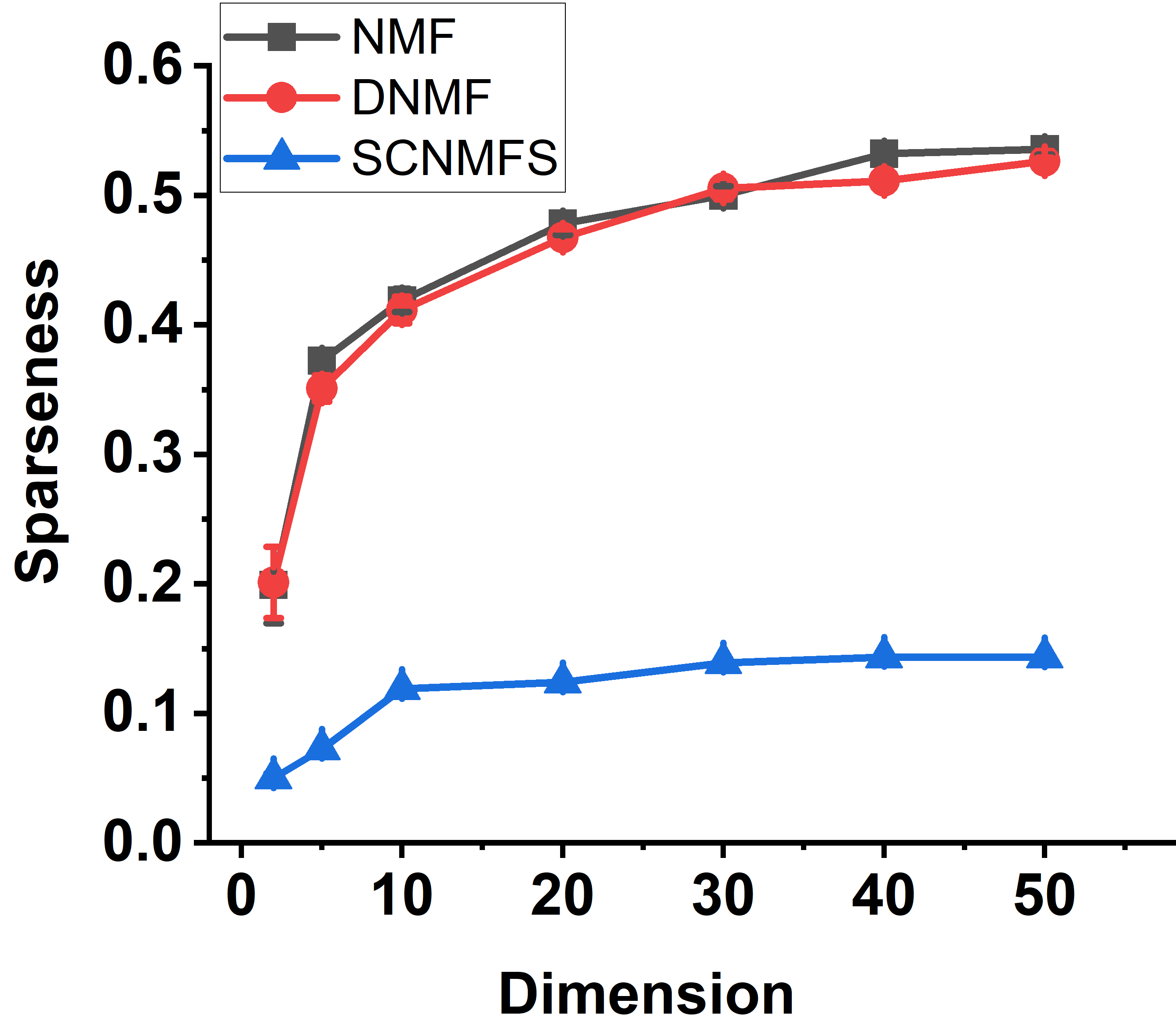}
}%
\subfigure[Subspace representations $\boldsymbol{V}_{\rm train}^{\Delta}$.]{
\centering
\label{fig:ROC:15a}
\includegraphics[width=0.48\linewidth]{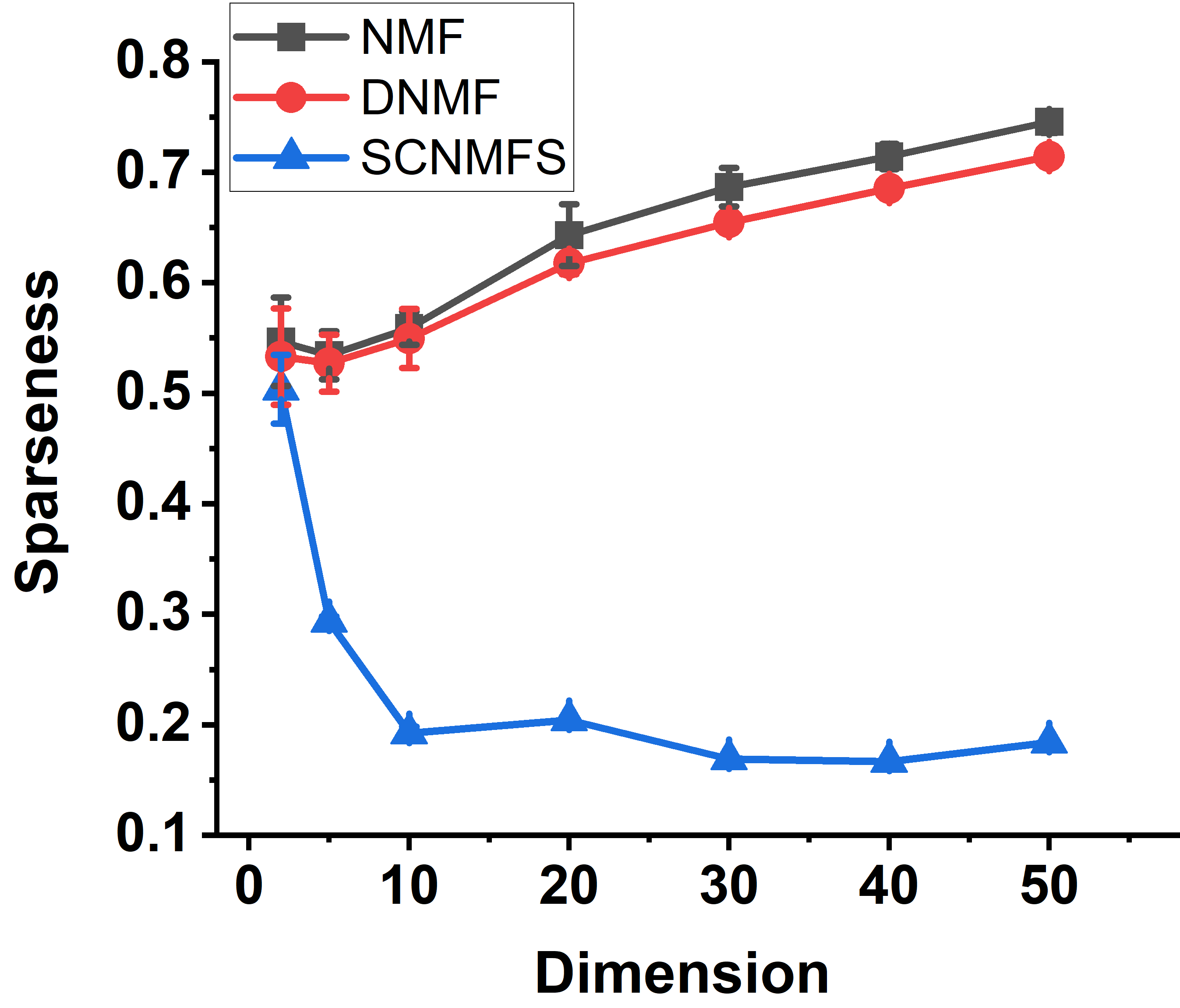}
}%
\caption{Sparsity analysis of NMF, DNMF and SCNMFS by investigating the projection matrices $\boldsymbol{U}_{\rm train}^{\Delta}$ and the subspace representations $\boldsymbol{V}_{\rm train}^{\Delta}$ with different subspace dimensions. Dataset {\tt PneumoniaMNIST} of size 300 is used in this experiment.} 
\label{fig:sparsity}
\end{figure}

\begin{figure}[htp]
\subfigure[Projection matrix $\boldsymbol{U}_{\rm train}^{\Delta}$]{
\label{fig:5-1}
\centering
\includegraphics[width=0.49\linewidth]{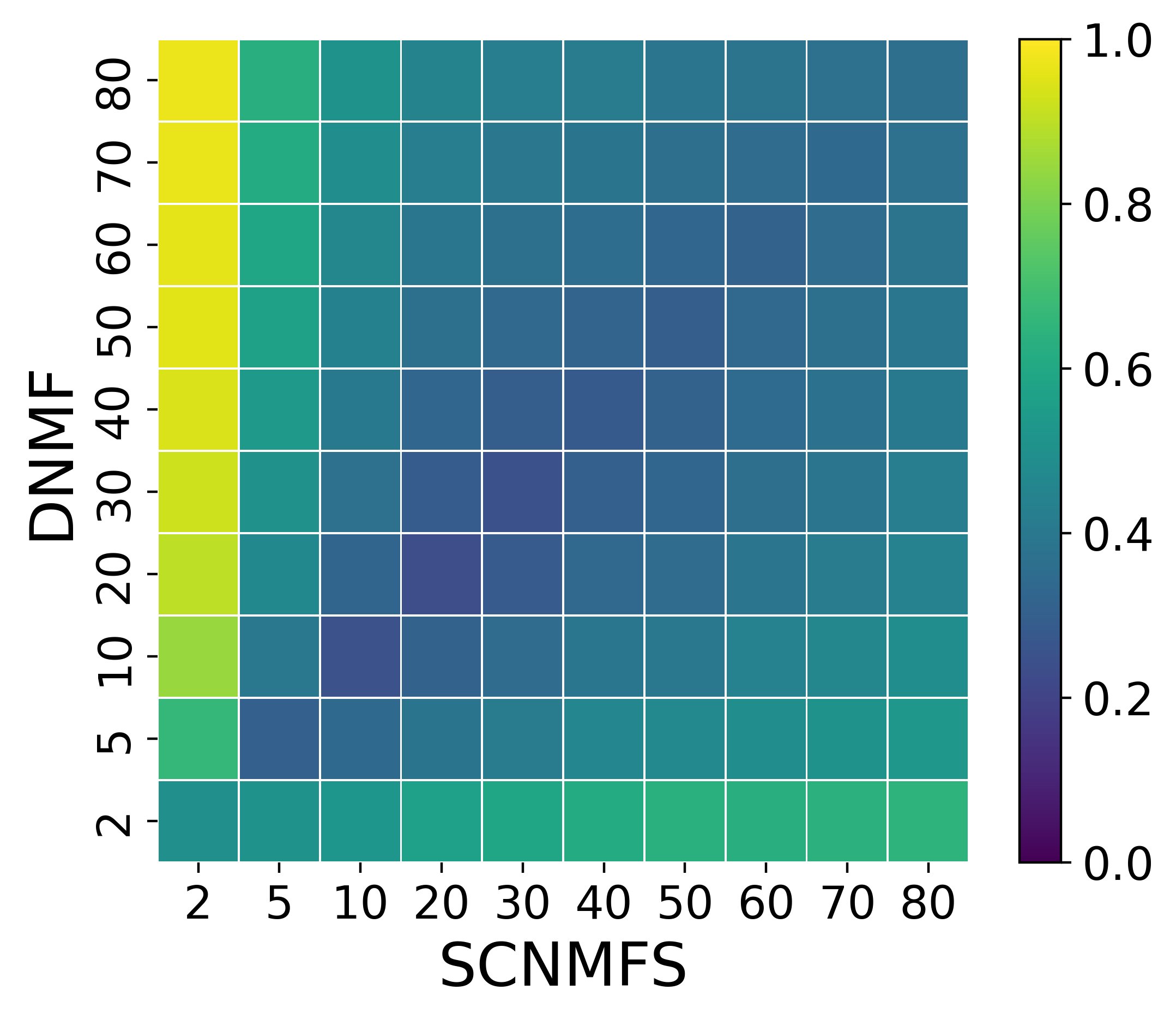}
}%
\subfigure[Subspace representation $\boldsymbol{V}_{\rm train}^{\Delta}$]{
\label{fig:5-2}
\centering
\includegraphics[width=0.49\linewidth]{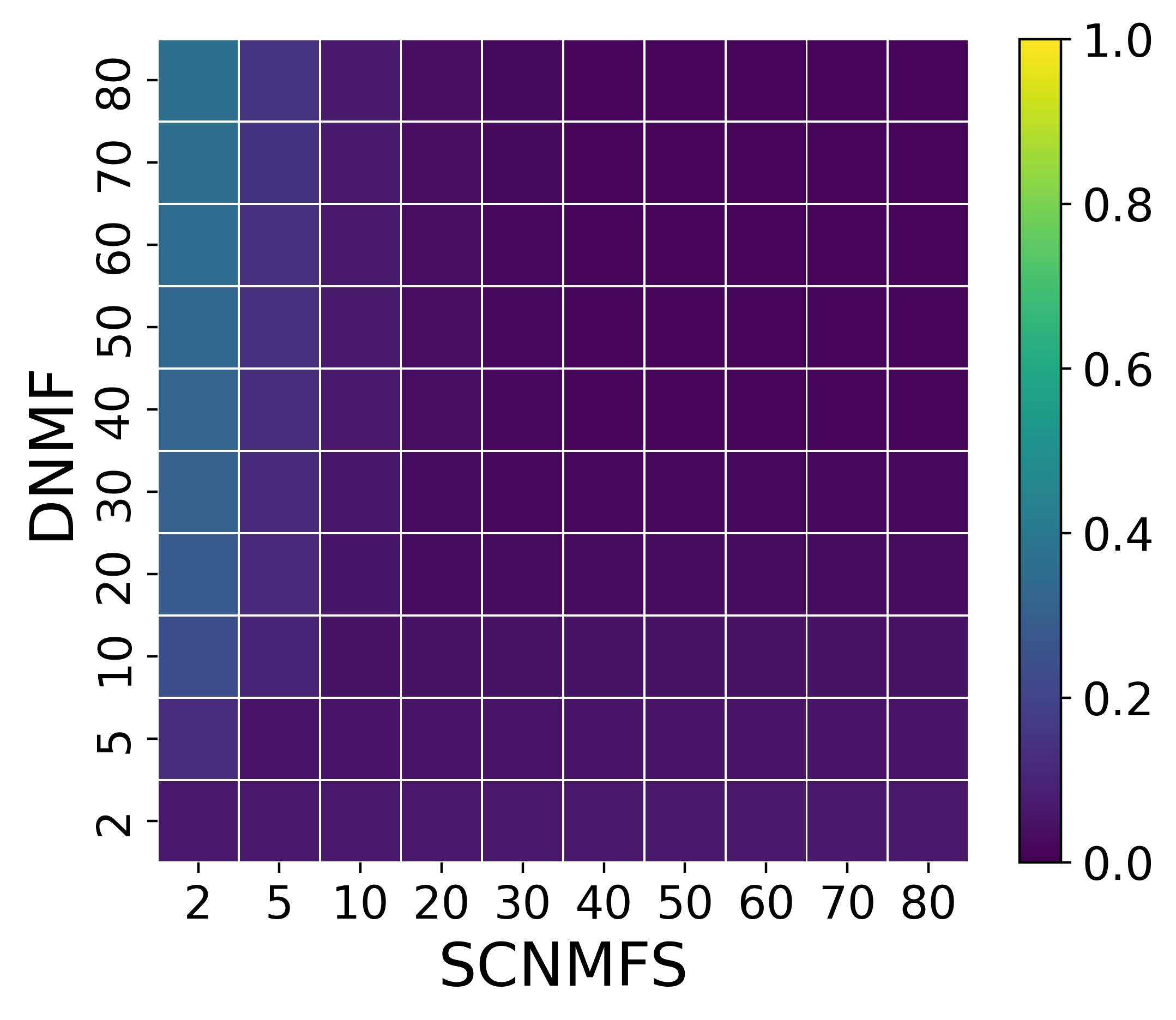}
}%
\caption{Correlation analysis regarding projection matrices $\boldsymbol{U}_{\rm train}^{\Delta}$ and subspace representations $\boldsymbol{V}_{\rm train}^{\Delta}$ generated by DNMF and SCNMFS with different subspace dimensions. Dataset {\tt PneumoniaMNIST} is used.} 
\label{fig:5}
\vspace{-1.2em}
\end{figure}

Fig. \ref{fig:sparsity} presents a further comparison between NMF and the supervised NMF methods in terms of sparsity of the generated subspaces. In detail, the sparsity comparison is conducted by investigating the projection matrix $\boldsymbol{U}_{\rm train}^{\Delta}$ and the subspace representation $\boldsymbol{V}_{\rm train}^{\Delta}$ generated by NMF, DNMF and SCNMFS on the {\tt PneumoniaMNIST} dataset. Note that compared to the standard NMF, which can generate sparse representation in subspace, DNMF and SCNMFS are different from it due to the way of combining labels. Fig. \ref{fig:sparsity} shows that the subspace generated by SCNMFS is less sparse. 
This occurs because SCNMFS consolidates data with identical labels into a single point (signifying that all elements with the same label in the subspace exhibit roughly equal activity), while DNMF organizes data from each class into distinct clusters along the axis. The limited correlation regarding the matrices $\boldsymbol{U}_{\rm train}^{\Delta}$ and $\boldsymbol{V}_{\rm train}^{\Delta}$ produced by these methods further reinforces this trend, as depicted in Fig. \ref{fig:5}.

\subsubsection{Discussion}
The results reported above support the initial finding that subspace-based few-shot learning is a promising direction in data-driven few-shot learning in medical imaging. More importantly, NMF and its variations can be a viable alternative to SVD. The basic vectors in the projection matrix of NMF preserve more suitable information than the principal components in SVD. 
For the few-shot learning model, e.g. the  prototypical network \cite{snell2017prototypical}, the effect is not as obvious as expected in medical imaging. Moreover, data augmentation could even introduce negative effect in medical imaging. 

For generally hundreds of data that are available in medical scenarios, the supervised NMF methods (e.g., DNMF and SCNMFS) are superior to the unsupervised subspace methods (e.g., SVD and NMF) in classification tasks, and the results of SCNMFS are more robust than DNMF in different subspace dimensions. 
Besides, we found that, from the data characteristics point of view, the non-negative representation in NMF will bring sparsity and enhance the ability of data representation in subspaces. Appropriate label signal combination (e.g., DNMF and SCNMFS) will further boost the discriminative ability of the subspace, which is particularly important for classification tasks with limited data.

Another feature of NMF is the part-based representation in the subspace. The non-negativity constraints are compatible with the intuitive notion of combining parts to form a whole. In the next subsection we further show how part-based representations of NMF can be improved by the supervised NMF methods via locating the discriminative image regions in medical images.

\subsection{Validation by Class Activation Map and Discussion}\label{subsec:Signal Information Analysis}
Finally, to further understand/validate the properties of the above tested different subspace methods, we develop a CAM method, see Appendix \ref{appendix:1} for the detail, based on the one proposed in \cite{zhou2016learning} to illustrate the discriminative image regions these subspace methods generated.

\begin{figure*}[!th]
\centering
\includegraphics[width=6.90in]{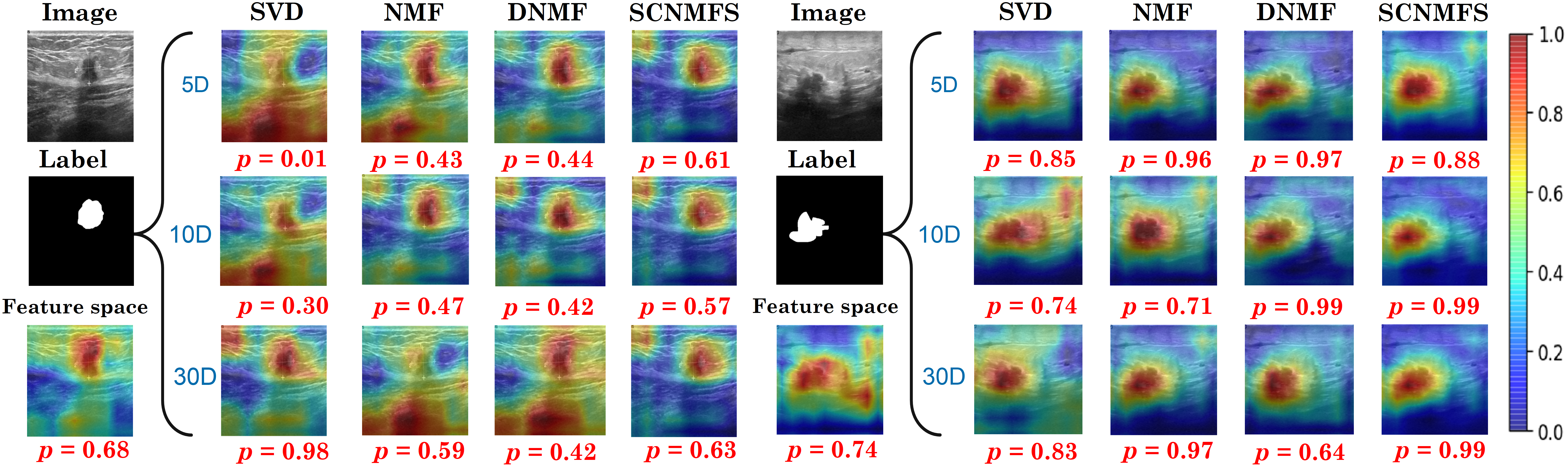}
\caption{
CAM validation for different subspace methods (i.e., SVD, NMF, DNMF and SCNMFS) on the breast ultrasound image dataset. Results of each method for 3 different dimensions (i.e., 5D, 10D and 15D) are considered. 
The discriminant regions, i.e., those regions to which the output decision is most sensitive to, appear as red.
Results under `Feature space' are the ones obtained without dimensionality reduction.  Predicted cancer class scores by each method are shown below the individual CAMs; particularly, $p > 0.5$ indicates the prediction is correct. Compared with the feature space and other subspaces, we see that SCNMFS achieves the best in positioning the lesion area and this localisation of discriminative features is consistent across dimensions. 
} 
\label{fig:CAM2}
\vspace{-1.0em}
\end{figure*}

Fig. \ref{fig:CAM2} gives the CAM results for different subspace methods (i.e., SVD, NMF, DNMF and SCNMFS) on the breast ultrasound image dataset \cite{breastmnist} with a training set size of $300$.
In Fig. \ref{fig:CAM2}, the result for feature space in the first column of each panel is the one obtained without dimensionality reduction, while the rest of the columns in each panel are the results of the subspace representations in 3 selected dimensions (i.e., $5$D, $10$D and $15$D for different rows in Fig. \ref{fig:CAM2}). The discriminant areas are highlighted in red. The probability score under each CAM is the predication result (to be cancer) of each subspace method.

It is observed that when the prediction is  correct, there is no significant disparity in the CAMs between the results of the subspaces  and the feature space, but the discriminant regions inferred from the subspaces are more centralised than the feature space, indicating the virtue of exploiting subspaces. For the incorrect predictions, the CAMs by NMF vary as the dimension increases, and the same tendency is also appeared in the CAMs by DNMF, see the second and third columns in the left panel in Fig \ref{fig:CAM2}. This demonstrates that although DNMF is a supervised variation of NMF, it does not fundamentally change the prediction performance compared to NMF. In contrast, the supervised NMF method SCNMFS performs excellently, i.e., the discriminative regions in its CAMs are stable and invariant to different dimensions.
In contrast, SVD suffers from incorrect prediction from its eigenvectors; moreover, as shown in Fig. \ref{fig:CAM2}, the discriminant regions in its CAMs shift as the dimension increases, indicating that the key information it uses for classification prediction is preserved in its eigenvectors, including the ones corresponding to small eigenvalues. This tendency is more obvious when the model with SVD makes incorrect predictions (i.e., the CAMs with $p < 0.5$); see the left panel in Fig. \ref{fig:CAM2}.

The results from the subspaces experiments empirically demonstrate that the part-based representation can be enhanced in supervised NMF, thereby contributing to lesion detection. The vector quantization property in SCNMFS ensures stability across different dimensions. Even with limited data, the discriminant areas achieved by SCNMFS closely align with the ground truth compared to other methods. Furthermore, the localized diagnosis by the CAM method underscores the effectiveness of the part-based representation provided by supervised NMF in identifying discriminative information in medical imaging.

\section{Conclusion}\label{sec:conclusion}

In this paper, we explored the innovative application of NMF (non-negative matrix factorization) and its supervised variations (i.e., DNMF and SCNMFS) as tools to gain insights regarding the properties of subspace features extracted from deep neural networks pre-trained on large natural image datasets, adapted for medical imaging in few-shot learning.
In our experiments, we found that for few-shot learning with limited datasets (e.g., a few hundred images), data augmentation methods are not as useful as widely believed, while reducing dimensionality could bring a suitable solution in the case that the magnitude of the data is smaller than the feature dimension.
Using this insight, we introduced NMF and supervised NMF as alternatives to replace the common dimension reduction method PCA/SVD. Our suggestion will alleviate the limitations of using PCA/SVD in multimodal data and shed lights on the link between non-negativity and sparsity in few-shot learning. 

By carrying out the experiments on $14$ medical datasets including $11$ distinct disease and 4 image acquisition modalities, our work exposes considerable additional fundamental and surprising findings as follows. 
I) Dimensionality reduction yields a constant performance advantage in the few-shot learning regime. 
II) The NMF-based representation, including its supervised variants (i.e., DNMF and SCNMFS), serves as a feasible alternative to SVD-based subspaces. SVD suffers from incorrect predictions from its eigenvectors for classification tasks with insufficient data. Moreover, utilizing the SCNMFS subspace instead of PCA/SVD-based variance-preserving dimensionality reduction yields significant performance improvements, even at extremely low dimensions.
III) The combination of label information in supervised NMF greatly impacts the interpretation of the subspace, e.g. SCNMFS gives a way of constructing a discriminative subspace with vector quantization rather than preserving the distribution as DNMF does. 
IV) The subspace sparsity does not considerably improve classification performance due to data scarcity and may even decrease the accuracy in the few-shot learning framework, yet non-negativity is more worthwhile to pursue given its part-based representation capacity. 
V) Part-based representations obtained from NMF-based subspaces can facilitate localization diagnosis in medical imaging, especially with limited medical images.

In this study, we adopted two supervised NMF methods, i.e., DNMF and SCNMFS, to investigate the effect of label information in generating discriminative subspaces. We acknowledge the significant contributions of NMF and its supervised variants, which encourage us to approach the concept of subspace from a different perspective. This also drives our investigation into the possibilities of a novel data-based few-shot learning approach, leveraging the features preserved within the NMF subspace for few-shot learning in medical imaging.
During our study, we encountered several limitations that warrant consideration. In addition to choosing a modest subspace dimension, our experiments focused solely on exploring subspaces using image data. However, in medical scenarios, multi-modal data (such as images and text) are often associated with the same disease. Future research could explore integrating multi-modal data into the analysis. This could involve investigating novel NMF variants and exploring alternative loss functions to enhance the stability and effectiveness of subspace representation in medical imaging analysis, thereby improving classification performance.

\section{Acknowledgments}
MN's contribution to this work is funded by Grant EP/S000356/1, Artificial and Augmented Intelligence for Automated Scientific Discovery, Engineering and Physical Sciences Research Council (EPSRC), UK.

\appendix
\section{Class Activation Map Generation}\label{appendix:1}

\begin{algorithm}[htb] 
\caption{CAM generation with subspaces} 
\label{alg:Framwork-heatmap} 
\begin{algorithmic}[1] 
\REQUIRE  $\boldsymbol{d}$, $f_{\Theta_1}$, $ f_{\Theta_2}$ and $\boldsymbol{U}_{\rm train}^{\Delta} $\\
\ENSURE Class activation map $\boldsymbol{R}$ 
\algrule
\STATE Obtain the feature map $\boldsymbol{M}$ and form $\bar{\boldsymbol{M}}$;
\STATE Obtain $ f_{\Theta_2} $ weight matrix $\boldsymbol{W} = (\boldsymbol{w}_1, \cdots, \boldsymbol{w}_C) \in \mathbb{R}^{k\times C}$
\STATE Obtain the predict label $i$ for image $\boldsymbol{d}$ from $f_{\Theta_2}$ and the corresponding vector $\boldsymbol{w}_i$;
\STATE Compute the weight vector $\boldsymbol{x}^{\prime}=\boldsymbol{U}_{\rm train}^{\Delta} \boldsymbol{w}_i$;
\STATE Form $\boldsymbol{R}^{\prime} \in \mathbb{R}^{h\times w}$ by reshaping $(\boldsymbol{x}^{\prime})^\top \bar{\boldsymbol{M}} \in \mathbb{R}^{h\cdot w}$;
\STATE Resize $\boldsymbol{R}^{\prime}$ to the size of the test image and generate $\boldsymbol{R}$.
\end{algorithmic}
\end{algorithm}
\vspace{-0.6em}

This Appendix shows the steps of calculating CAMs for our studied few-shot learning framework in medical imaging. 
%
Let $\boldsymbol{M} \in \mathbb{R}^{ c \times h \times w}$ be the feature map produced by the last convolutional layer of the pre-trained network $f_{\Theta_1}$ corresponding to the input image $\boldsymbol{d} \in \boldsymbol{\cal D}$,
where $c$ represents the number of channels, and $h$, $w$ are the size of the feature map in each channel. We flatten the channel-wise feature maps, and then $\boldsymbol{M}$ is changed into a matrix $\bar{\boldsymbol{M}}$, $\bar{\boldsymbol{M}} \in \mathbb{R}^{ c \times 	\left(h \cdot w\right)}$.
The pre-trained model $ f_{\Theta_1}$ transfers the input image $\boldsymbol{d}$ into a vector $\boldsymbol{x} \in \mathbb{R}^{c}$ in the feature space, i.e., $\boldsymbol{x} \leftarrow { f_{\Theta_1}(\boldsymbol{d})}$. 
With the selected rank $k$ and subspace method $\Delta$, the corresponding subspace representation $\boldsymbol{v}^{\Delta} \in \mathbb{R}^{k}$ is generated by using the projection matrix $\boldsymbol{U}_{\rm train}^{\Delta} \in \mathbb{R}^{c \times k}$. 
Let  $\boldsymbol{W} = (\boldsymbol{w}_1, \cdots, \boldsymbol{w}_C) \in \mathbb{R}^{k\times C}$ be the weight matrix of $f_{\Theta_2}$. 
Since $f_{\Theta_2}$ is a fully connected layer, the predicted process for the input image $\boldsymbol{d}$ can be regarded as $\boldsymbol{z} \approx (\boldsymbol{v}^{\Delta})^\top\boldsymbol{W}$. Assume class $i$ is predicted (i.e., $\boldsymbol{z}$'s largest component is the $i$-th entry) for image $\boldsymbol{d}$. Then the weight vector $\boldsymbol{w}_i$ is used to generate $\boldsymbol{x}^{\prime}=\boldsymbol{U}_{\rm train}^{\Delta} \boldsymbol{w}_i \in \mathbb{R}^{c}$, which will be used as a weight vector for $\bar{\boldsymbol{M}}$. 
The initial CAM result is formed by reshaping $(\boldsymbol{x}^{\prime})^\top \bar{\boldsymbol{M}} \in \mathbb{R}^{h\cdot w}$ to a matrix $\boldsymbol{R}^{\prime}$, $\boldsymbol{R}^{\prime} \in \mathbb{R}^{h \times w}$.
The final CAM result $\boldsymbol{R}$ is drawn by resizing $\boldsymbol{R}^{\prime}$ to the size of the test image by interpolation.
The above steps are summarised in Algorithm \ref{alg:Framwork-heatmap}.

\bibliographystyle{IEEEtran}
\bibliography{cas-refs}

\end{document}